\documentclass{article} % For LaTeX2e
\usepackage{iclr2026_conference,times}
\usepackage{amsmath,amsfonts,bm}

\def\eqref#1{equation~\ref{#1}}
\def\1{\bm{1}}

\DeclareMathAlphabet{\mathsfit}{\encodingdefault}{\sfdefault}{m}{sl}
\SetMathAlphabet{\mathsfit}{bold}{\encodingdefault}{\sfdefault}{bx}{n}

\usepackage{hyperref}
\usepackage{url}
\usepackage{graphicx}
\usepackage{booktabs}  % 用于 \toprule, \midrule, \bottomrule
\usepackage{multirow}  % 用于 \multirow
\usepackage{makecell}  % 用于 \makecell
\usepackage{tabularx}
\usepackage{longtable}
\usepackage{colortbl}
\usepackage{pifont} 
\usepackage{subcaption}
\usepackage{caption}
\usepackage[percent]{overpic}
\usepackage{adjustbox}
\usepackage{listings}
\usepackage{upquote}
\usepackage{xcolor}
\makeatletter
\renewcommand\@makefntext[1]{\noindent\makebox[1.8em][r]{}#1}
\makeatother
\newcommand{\cmark}{\ding{51}} 
\newcommand{\xmark}{\ding{55}}
\definecolor{darkpink}{rgb}{0.91, 0.33, 0.5}
\definecolor{mygray}{gray}{.94}

\title{TOUCH: Text-guided Controllable Generation of Free-Form Hand-Object Interactions}

\author{Guangyi Han$^{1}$, Wei Zhai$^{1,\dagger}$, Yuhang Yang$^{1}$, Yang Cao$^{1}$, Zheng-Jun Zha$^{1}$\\
{$^{1}$~University of Science and Technology of China} \\
\small{\texttt{\{hanguangyi@mail., wzhai056@, yyuhang@mail., forrest@, zhazj@\}ustc.edu.cn}}
}
\iclrfinalcopy % Uncomment for camera-ready version, but NOT for submission.
\newcommand{\methodname}{TOUCH}
\newcommand{\dstname}{WildO2}

\begin{document}
\maketitle
\begin{abstract}
Hand-object interaction (HOI) is fundamental for humans to express intent. Existing HOI generation research is predominantly confined to fixed grasping patterns, where control is tied to physical priors such as force closure or generic intent instructions, even when expressed through elaborate language. Such an overly general conditioning imposes a strong inductive bias for stable grasps, thus failing to capture the diversity of daily HOI. To address these limitations, we introduce \textbf{Free-Form HOI Generation}, which aims to generate controllable, diverse, and physically plausible HOI conditioned on fine-grained intent, extending HOI from grasping to free-form interactions, like pushing, poking, and rotating. To support this task, we construct \textbf{\dstname}, an in-the-wild diverse 3D HOI dataset, which includes diverse HOI derived from internet videos. Specifically, it contains 4.4k unique interactions across 92 intents and 610 object categories, each with detailed semantic annotations. Building on this dataset, we propose \textbf{\methodname}, a three-stage framework centered on a multi-level diffusion model that facilitates fine-grained semantic control to generate versatile hand poses beyond grasping priors. This process leverages explicit contact modeling for conditioning and is subsequently refined with contact consistency and physical constraints to ensure realism. Comprehensive experiments demonstrate our method's ability to generate controllable, diverse, and physically plausible hand interactions representative of daily activities. The project page is \href{https://guangyid.github.io/hoi123touch}{here}. 
\end{abstract}
\footnotetext{$\dagger$ Corresponding Author.}
\section{Introduction}
Hand-Object Interaction (HOI) is fundamental to expressing intent and executing tasks in human daily life, 
and the ability to generate controllable interactions is crucial for AR/VR, robotics, and embodied AI \citep{zheng2025survey}.
While existing HOI generation research has progressed from ensuring physical plausibility \citep{fang2020graspnet} to incorporating semantic controllability \citep{li2024semgrasp,Yang_2023_ICCV}, 
its scope remains confined to a grasp-centric paradigm \citep{zhou2022toch, zhou2024gears}, where control signals—spanning from physical constraints like force closure \citep{nguyen1988constructing, zheng2005coping} 
to coarse instructions like verb-noun pairs \citep{li2025latenthoi}—are overly general, 
imposing a strong inductive bias that favors the generation of stable grasps \citep{grab,turpin2022grasp}, sacrificing interaction diversity.
Furthermore, even with more sophisticated control such as detailed natural language (e.g., via LLMs) \citep{huang_etal_cvpr25, zhang2025openhoi, zhang2025pear,shao2025great}, 
the underlying model designs and inherent inductive biases are still fundamentally geared towards generating only grasping interactions, driven by historical focus and prevailing representations. 
Consequently, these approaches lack the fine-grained control and inherent capability to capture the diverse non-grasping interactions found in the real world, including varied hand poses, contact details, and nuanced semantic intent.

To bridge the gap between the limited scope of current methods and the complexity of real-world interactions, we introduce the task of \textbf{Free-Form HOI Generation}. 
The goal is to break grasp-centric limitations and shift towards generating diverse interactions, including the vast array of non-grasping manipulations. This task emphasizes expressiveness and controllability in the generation process, aiming to synthesize interactions that are not only physically plausible but also semantically rich and truly adaptable to complex human intentions.
% It paves the way for next-generation HOI systems capable of realizing complex intentions with high fidelity, moving beyond predefined interaction patterns

The core challenge of this task lies in two aspects: what to generate and how to generate it. 
The former pertains to spatial plausibility: the model must break free from restrictive grasping priors 
(e.g., palm position and orientation, contact region assumption \citep{ye2023ghop,jiang2021graspTTA,jung2025haco}) to explore a vast yet physically valid interaction space. 
To address this, we propose that contact relationships serve as a powerful cue to constrain this high-dimensional space, offering a more nuanced understanding of physically valid interactions.
The latter pertains to semantic controllability: the model must accurately map fine-grained textual instructions to specific hand configurations and contact regions. 
The prior knowledge within Large Language Models (LLMs) offers a promising pathway for this guidance \citep{tang2023graspgpt}. 
A major obstacle is the lack of diverse 3D training data; current datasets\citep{zhan2024oakink2,fu2025gigahands} are limited to lab-based grasps, and large-scale real-world data collection remains challenging.
In contrast, abundant 2D HOI videos online provide rich and realistic daily interaction behaviors.

To tackle the proposed task and challenges, we present \methodname, a three-stage framework for controllable free-form HOI generation.
First, we explicitly model the contact on the surfaces of the hand and the object separately by jointly encoding spatial point-cloud relations and semantic information, 
providing strong spatial priors to mitigate uncertainty from the high degrees of freedom in interaction position and pose. We further incorporate part-level hand modeling for more precise action control.
Second, we employ a multi-level diffusion model with attention-based fusion of semantics and geometry: coarse-grained intent and global object geometry guide the early diffusion stages, 
while fine-grained text and local contact features refine detailed motions in deeper stages, enabling fine-grained semantic controllability.
Finally, we introduce self-supervised contact consistency and physical plausibility constraints to optimize the generated interactions, ensuring realism and physical feasibility.
Compared to prior methods restricted to grasp generation, \methodname{} naturally generalizes to diverse free-form HOI such as pushing, pressing, and rotating.

Additionally, based on 3D object reconstruction \citep{instantmesh}, we introduce an automated pipeline to build the dataset \dstname{} that jointly recovers and optimizes high-quality 3D hand-object interaction samples from internet videos annotated with interaction intent. By leveraging vision-language models \citep{Qwen-VL}, we generate fine-grained semantic annotations, resulting in the 3D daily HOI dataset covering diverse interaction intents. 
% This dataset supports the training and evaluation of our task and provides diversity in hand poses generated by our model.

\begin{figure}[t]
    \begin{center}
    \includegraphics[width=1\textwidth]{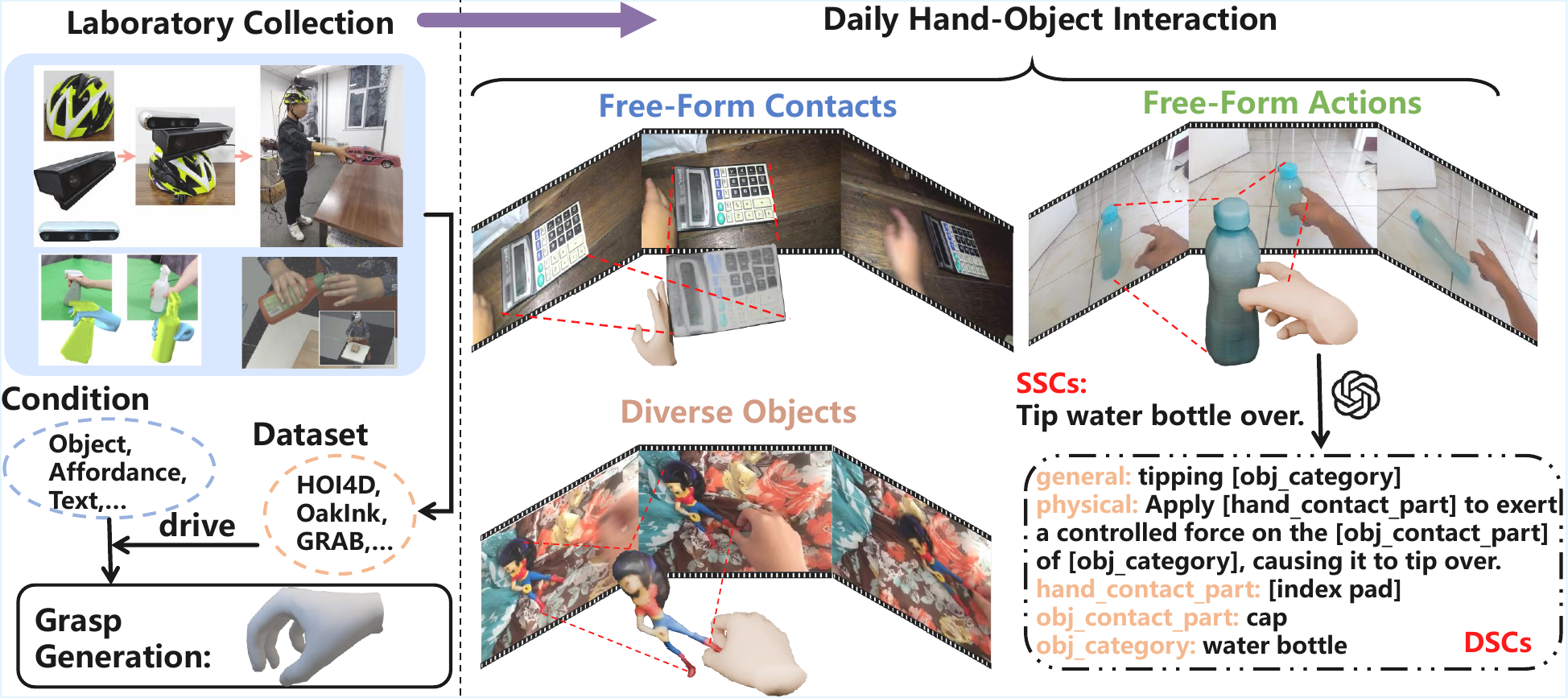}
    \end{center}
    \caption{\textbf{Overview.} We extend HOI generation beyond laboratory “grasp” settings (left) toward broader daily HOI modalities (right), enabling the modeling of more human-like interactions. Our dataset \dstname, built from Internet videos, covers more contacts, more objects, and more actions, and is enriched with descriptive synthetic captions (DSCs) to support fine-grained semantic controllable HOI generation with our method, \methodname.}
    \label{fig:overview}
    \vspace{-2em}
\end{figure}
Our main contributions are:
(1) We propose to extend the HOI from constrained grasping to a broader, more realistic, and more diverse set of daily interactions.
(2) We propose \methodname{}, a new framework that can generate natural, physically reasonable, and diverse free-form HOI under fine-grained text guidance.
(3) We build an automated pipeline and construct \dstname{}, an in-the-wild 3D dataset for daily HOI, providing a critical resource that enables future research in this domain.
Extensive experiments demonstrate the superiority of \methodname{}.
\section{RELATED WORK}
\subsection{Hand-Object Interaction Datasets.}
Existing 3D hand-object interaction (HOI) datasets are predominantly collected in controlled laboratory settings, 
relying either on physics-based simulation synthesis \citep{obman} or motion capture systems to record real interactions \citep{ho3d,hoi4d,oakink,ctcpose}. 
Although these datasets provide valuable support for modeling 3D HOI, they suffer from limited diversity due to constrained camera setups, 
a small number of participants, and a restricted set of object instances. 
In contrast, large-scale in-the-wild video datasets \citep{epickitchens,ego4d,100doh} contain abundant HOI clips, but lack high-quality 3D annotations. 
Some studies have attempted to annotate subsets of these videos in 3D using object template-based optimization methods \citep{MOW,RHOV}; 
however, due to the high diversity of open-set objects, scaling such approaches remains challenging.

\subsection{Template-Free HOI Reconstruction.}
The core bottleneck in reconstructing HOIs in the wild has long been the recovery of diverse object geometries. 
While existing template-free approaches \citep{fan2024hold, ye2022s} avoid predefined object model constraints, 
they are typically trained on limited datasets and exhibit poor generalization to novel objects.
In recent years, multi-view diffusion models \citep{zero123} and large-scale reconstruction models (LRMs) \citep{hong2024lrm} 
have enabled high-quality 3D mesh reconstruction directly from single images \citep{instantmesh} or text prompts \citep{poole2022dreamfusion}, demonstrating strong generalization capabilities. 
Motivated by these advances, several HOI studies have explored image-to-3D reconstruction pipelines to handle open-set objects in the wild. 
However, due to severe hand occlusion, these methods often rely on image inpainting to complete occluded regions \citep{funhoi,easyhoi,human3dhoi}, 
or employ text-to-3D generation to align with coarse reconstruction results \citep{mccho,web2grasp}. 
Nonetheless, most of these pipelines depend on heuristic completion or registration strategies, resulting in limited geometric consistency with the input, 
and have yet to be validated at scale in an automated manner.

\subsection{Data-driven Controllable HOI Generation.}
In the evolution of HOI generation, interaction guidance has progressively advanced: from coarse control based on grasp type \citep{feix2015grasp,chen2025dexonomy}, 
to object-conditioned generation \citep{graspingfield}, and further to task/action-level intent constraints \citep{christen2024diffh2o,yu2025hero,yang2024egochoir,yang2024lemon}. 
To enhance physical plausibility, contact penetration loss and hand anatomical constraints have been widely adopted \citep{wei2024grasp}. 
Additionally, explicit modeling of hand part segmentation and contact relationships with objects has been shown to improve physical realism and detailed expression of interactions \citep{contactgen,zhang2024nl2contact}. 
Building on these efforts, we propose a multi-level controllable generation framework trained on our newly constructed daily HOI dataset, 
enabling finer-grained semantic intent control and the flexible generation of free-form HOIs that align with complex human intentions.
\section{DATASET}

\begin{figure}[ht]
\begin{center}
\includegraphics[width=0.9\textwidth]{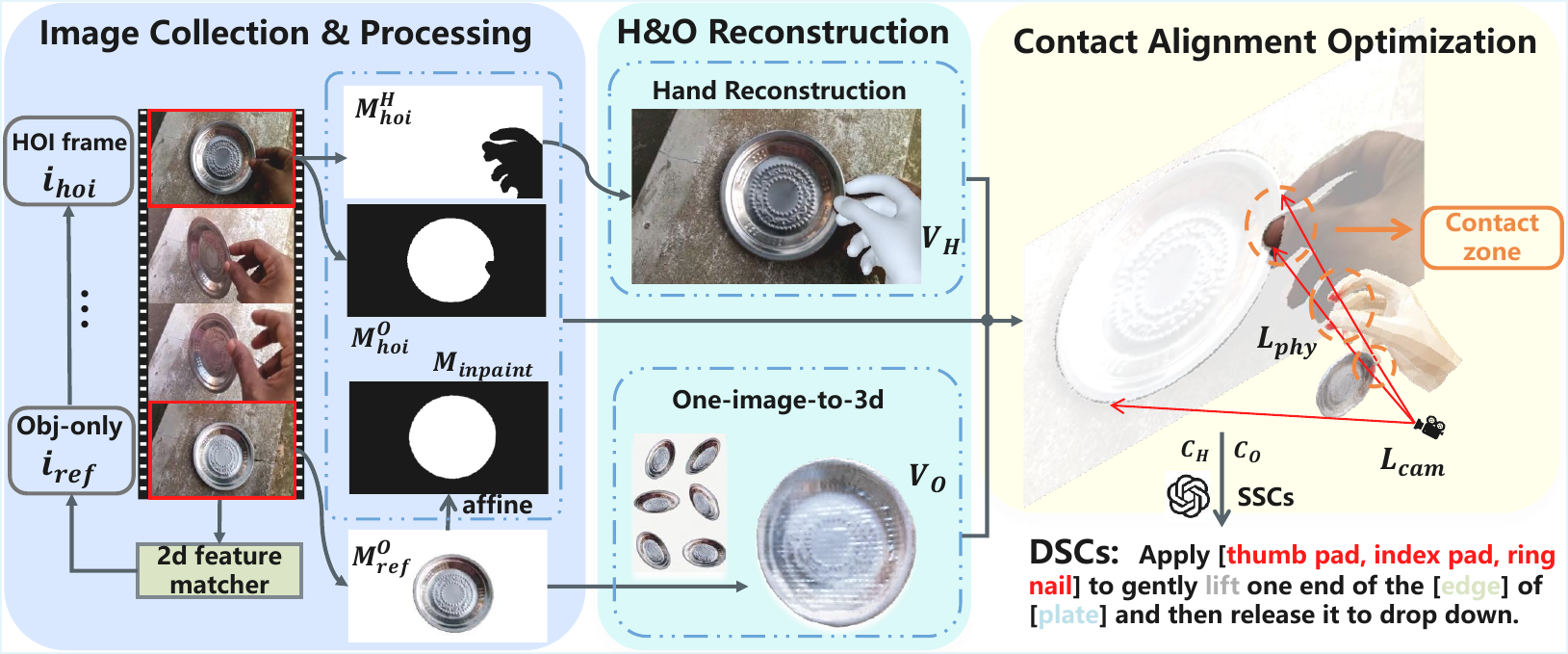}
\end{center}
\caption{The proposed data generation pipeline for \dstname{}. The process begins with O2HOI frame pair extraction from in-the-wild videos, followed by a three-stage pipeline for 3D reconstruction, camera alignment, and hand-object refinement that produces high-fidelity interaction data.}
\label{fig:dst_pipeline}
\vspace{-1em}
\end{figure}

\subsection{Data Collection and Processing}
Our goal is to construct a diverse dataset of 3D hand-object interactions from in-the-wild videos. A primary challenge in this process is the severe occlusion of the object by the hand, which compromises the quality of 3D object reconstruction. To address this, we introduce a semi-automated data generation pipeline, centered around a novel Object-only to Hand-Object Interaction (O2HOI) frame pairing strategy.\\
We begin by filtering the Something-Something V2 dataset \citep{sthv2}, which is rich in goal-directed human actions, to obtain 8k single-hand, single-object interaction clips. For each clip, we automatically extract an O2HOI pair (details in Appendix): an object-only frame $\mathbf{I}_{ref}$, where the object is unoccluded, and a corresponding interaction frame $\mathbf{I}_{hoi}$. To obtain a complete object mask in the interaction frame, we segment the object in $\mathbf{I}_{ref}$ using SAM2 \citep{ravi2024sam2} and then transfer this mask to $\mathbf{I}_{hoi}$ via a robust dense matching model \citep{edstedt2024roma}, yielding $\mathbf{M}_{{inpaint}}$. 
This mask transfer strategy offers a distinct advantage over common alternatives: 
it avoids the geometric inconsistencies of diffusion-based inpainting \citep{easyhoi} while being significantly more scalable than manual completion \citep{human3dhoi}. 
Consequently, our approach facilitates the automated, large-scale generation of high-fidelity 3D assets for reconstruction.

\subsection{Data Generation Pipeline}
Based on the O2HOI pairs, we build a three-stage pipeline to recover 3D HOI (see Figure~\ref{fig:dst_pipeline}).\\
\textbf{Stage 1: Initialization.} 
For each pair, we reconstruct a textured object mesh $\mathbf{V}^O_{{recon}}$ from the object-only frame $\mathbf{I}_{ref}$ using an image-to-3D model \citep{instantmesh}. 
Concurrently, we estimate initial MANO \citep{mano} hand parameters $\mathbf{H}_{{init}}$ from the interaction frame $\mathbf{I}_{hoi}$ using a state-of-the-art hand reconstruction method \citep{hamer}.\\
\textbf{Stage 2: Camera Alignment.} A challenge arises from coordinate system misalignment: 
the object mesh $\mathbf{V}^O_{{recon}}$ is created in a canonical space of the object-only frame $\mathbf{I}_{ref}$, 
while the hand exists in the camera space of the interaction frame $\mathbf{I}_{hoi}$. 
To unify them, we align $\mathbf{V}^O_{{recon}}$ to an object-centric global coordinate system relative to the interaction frame 
by optimizing the camera projection matrix $\mathbf{K}$ and extrinsics $(\mathbf{R}, \mathbf{t})$. This is achieved by minimizing a camera alignment loss, $L_{{cam}}$, via differentiable rendering. 
The optimization proceeds in two phases: we initially use mask IoU, Sinkhorn \citep{cuturi2013sinkhorn} loss, and an edge penalty term (to prevent the object from moving out of view). 
Once the IoU surpasses a threshold, we introduce scale-invariant depth \citep{eigen2014depth} and RGB reconstruction losses for fine-tuning.
The overall objective is formulated as:
\begin{equation}
\min_{\mathbf{K}, \mathbf{R}, \mathbf{t}} ; L_{{cam}} = L_{{mask}} + L_{{sinkhorn}} + L_{{edge}} + \lambda_{{fine}}(L_{{depth}} + L_{{rgb}}).
\label{eq:cam_align}
\end{equation}
\textbf{Stage 3: Hand-Object Refinement.} 
With the aligned camera and object, we refine the initial hand parameters $\mathbf{H}_{{init}}$ to achieve physically plausible contact. 
Specifically, we cast rays from the camera center through pixels within the interaction mask $\mathbf{M}_{{inpaint}}$. 
The intersection points of these rays with the 3D hand and object geometries define a potential 3D contact zone. 
We then optimize $\mathbf{H}$ using a refinement objective $L_{{align}}$, which combines 2D evidence with 3D physical constraints:
hand mask IoU ($L_{{mask}}^H$), 2D joint reprojection error ($L_{{j2d}}$), an ICP loss on the 3D contact zone ($L_{{icp}}$), and physical constraints for contact, penetration, and anatomy based on \citep{hasson2019learning_cpf}.
\begin{equation}
\min_{\mathbf{H}} ; L_{{align}} = L_{{mask}}^H + L_{{j2d}} + L_{{icp}} +  L_{{phy}}, \quad L_{{phy}} = L_{{contact}} + L_{{pene}} + L_{{anatomy}} + L_{{self}}.
\label{eq:hand_refine}
\end{equation}

This pipeline yields 4,414 high-quality 3D hand-object interaction samples after a final stage of manual inspection and refinement, which constitute the ground truth of our dataset.

\begin{figure}[ht]
    \centering
    \begin{minipage}{0.6\textwidth} 
        \centering
        \includegraphics[width=\linewidth]{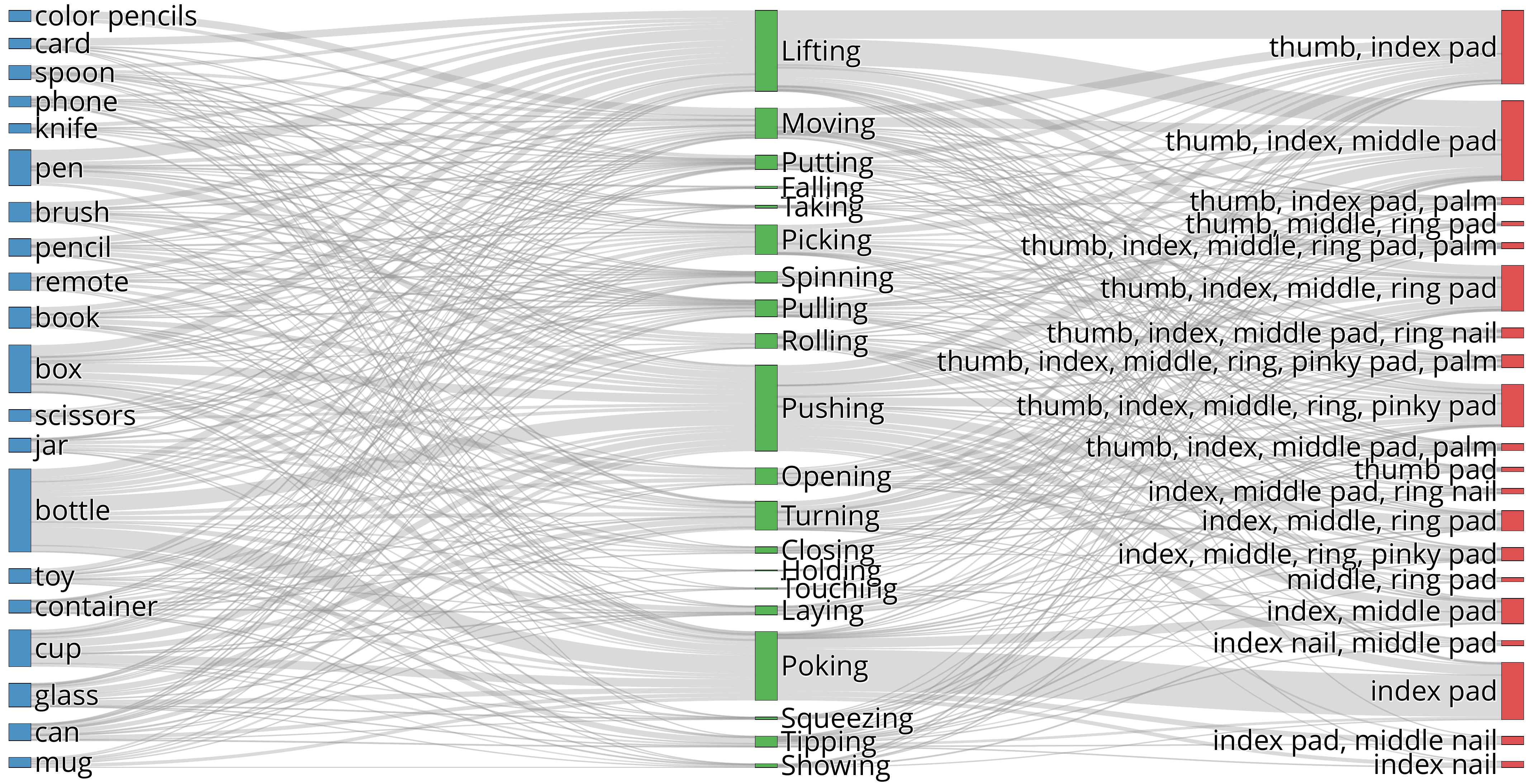}
        \subcaption*{(a)}
    \end{minipage}\hfill
    \begin{minipage}{0.38\textwidth} 
        \centering
        \includegraphics[width=\linewidth]{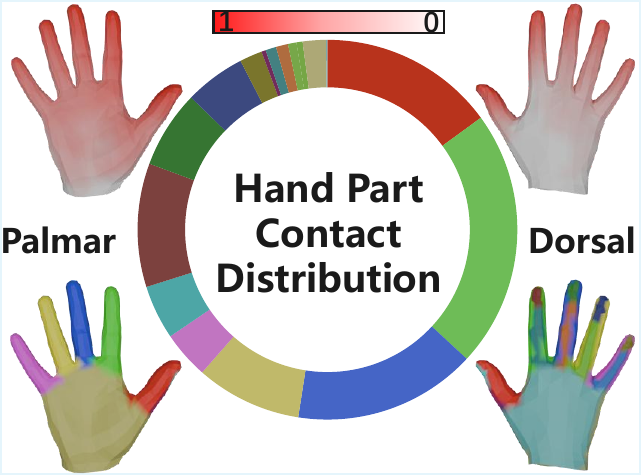}
        \subcaption*{(b)}
    \end{minipage}
  \caption{Dataset Distribution. (a) An illustration of the interplay between the most frequent object categories, interaction types, and hand contact regions. Object and action definitions are adapted and refined from \citep{sthv2}. Contact regions are derived based on our dataset analysis. (b) Specific segmentation of the 17 hand parts and their contact frequency distribution in the dataset, along with a contact heatmap of the entire hand.}
  \label{fig:dst_analysis}
   \vspace{-1em}
\end{figure}

\subsection{Data Annotation and Statistics}
We enrich our dataset with a multi-level annotation system, generating over 44k annotations. A statistical overview is provided in Figure~\ref{fig:dst_analysis}, with further details in the Appendix.\\
\textbf{3D Geometry and Transformation.} Each sample includes the final hand-object meshes $(\hat{\mathbf{V}_H}, \hat{\mathbf{V}_O})$ 
and the corresponding camera parameters derived from our generation pipeline.
\textbf{Contact Maps.} We compute dense contact maps between the hand and object surfaces. To handle varying object scales, 
our method robustly identifies contact regions by combining relative and absolute distance thresholds with bidirectional nearest-neighbor filtering.
\textbf{Multi-Level Language Descriptions.} We provide two levels of textual descriptions. 
We inherit the template-based Short Synthetic Captions (SSCs) from Something-Something V2 (e.g., "picking [Something] up"). 
Additionally, we use a Vision-Language Model (VLM) \citep{Qwen-VL} to generate more detailed Descriptive Synthetic Captions (DSCs), 
which are manually verified for quality and relevance.
\textbf{Fine-Grained Hand Part Segmentation.} We segment the hand mesh into 17 parts, including finger pads, nails, knuckles, palmar, and the dorsal region. 
This partitioning scheme goes beyond the coarse divisions commonly used in grasp generation tasks \citep{obman,contactgen}—which often focus only on contact on the inner hand—by also accounting for contact on the dorsal side. 
This fine-grained segmentation supports detailed local interaction analysis and facilitates alignment with the semantic descriptions in the DSCs.

\section{METHOD}
\begin{figure}[t]
    \begin{center}
    \includegraphics[width=1.0\textwidth]{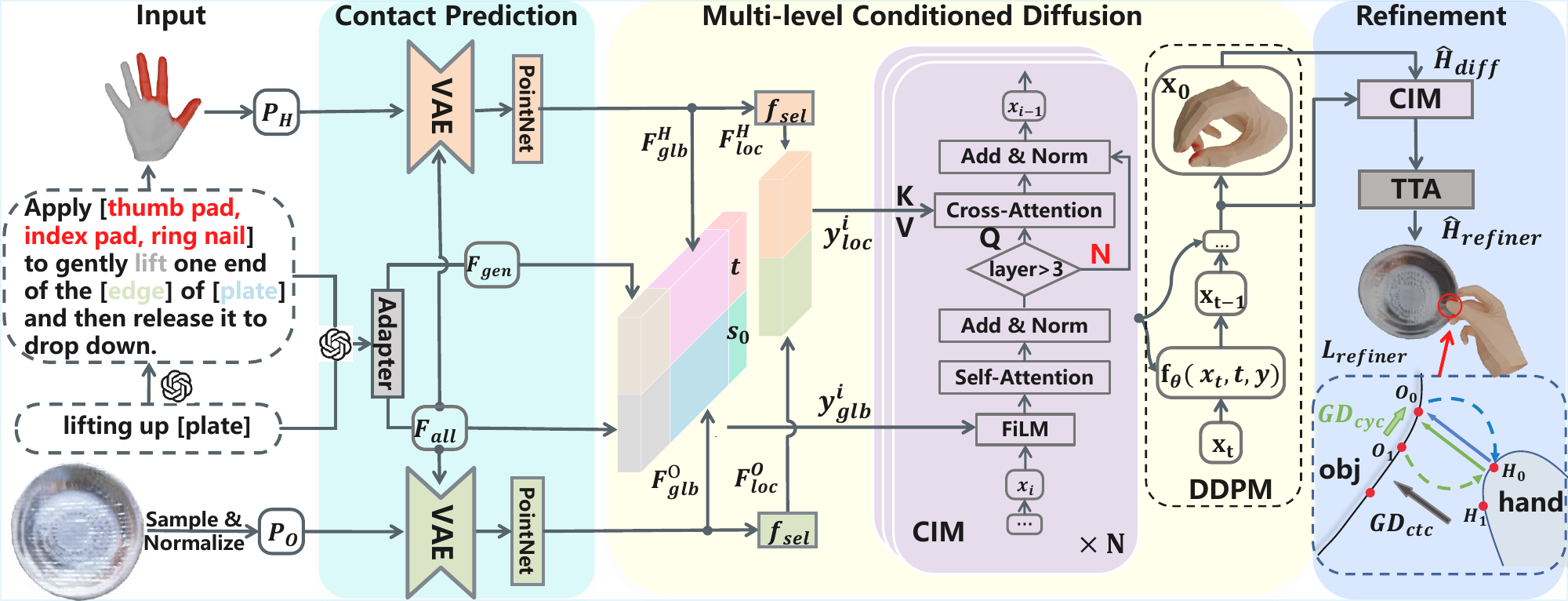}
    \end{center}
    \caption{Overview of our three-stage framework \methodname{} for generating hand-object interactions from multi-level text prompts and object meshes. 
    CIM stands for the Condition Injection Module.}
    \label{fig:method_pipeline}
    \vspace{-1em}
\end{figure}

This work aims to generate natural and physically plausible hand-object interaction (HOI) poses, parameterized by $\mathbf{H}$, 
along with corresponding contact maps $\mathbf{C}_H$ and $\mathbf{C}_O$, conditioned on a multi-level textual prompt $\mathbf{T}$ and an object mesh $\mathbf{V}_O$. 
To tackle this problem, we propose a three-stage framework, as illustrated in Figure~\ref{fig:method_pipeline}.
Specifically, the Contact Map Prediction module (Section~\ref{subsec:CMP}) infers the potential contact regions on the hand and object surfaces based on the text and object geometry. 
The Multi-Level Conditioned Diffusion module (Section~\ref{subsec:MLCD}) synthesizes a coarse hand pose 
by integrating coarse-to-fine textual and geometric features within a diffusion framework, ensuring alignment with multi-level constraints. 
Finally, the Physical Constraints Refinement module (Section~\ref{subsec:PCR}) further optimizes the coarse pose to enhance contact realism and prevent penetrations.

\subsection{Contact map prediction}
\label{subsec:CMP}
To generate diverse interactions beyond simple grasping, we design two independent yet similar CVAEs \citep{sohn2015learning} to generate binary contact maps for the object and the hand, respectively.\\
For the object branch, we sample a point cloud $\mathbf{P}_O \in \mathbb{R}^{N_O \times 3}$ ($N_O$ = 3000) from its mesh $\mathbf{V}_O$, normalize it, and record the scale factor $s_O$. We use PointNet \citep{qi2016pointnet} to extract its geometric features, which are concatenated with $s_O$ to form the object condition $\mathbf{F}_{O}$. For the hand branch, we generate a canonical point cloud $\mathbf{P}_H^0 \in \mathbb{R}^{N_H \times 3}$ ($N_H$ = 778) from MANO's zero pose and shape parameters $\mathbf{H}^0$. This point cloud, combined with a hand-part mask initialized from the fine-grained text $\mathbf{T}_{DSC}$, is processed by PointNet to obtain the hand condition $\mathbf{F}_H$. This design integrates the topological structure of the point clouds with text-guided emphasis on interaction-relevant hand regions. Both CVAEs are trained conditioned on their respective geometric features ($\mathbf{F}_O, \mathbf{F}_H$) and a shared text feature $\mathbf{F}_{DSC} = f_{text}(\mathbf{T}_{DSC})$, which is extracted using the Qwen-7B \citep{qwen} processed through a lightweight adapter. The optimization objective is a composite loss function:
\begin{equation} L_{contact} = L_{focal} + L_{dice} + \beta L_{KL} \end{equation}
where $L_{focal}$ and $L_{dice}$ supervise the contact prediction, and $L_{KL}$ structures the latent space. During inference, under the conditional features $(\mathbf{F}_{O}, \mathbf{F}_{H}, \mathbf{F}_{DSC})$, the model samples from a Gaussian prior $z \sim \mathcal{N}(0,I)$ and decodes it to produce the predicted binary contact maps $\hat{\mathbf{C}}_O \in \{0,1\}^{N_O \times 1}$ and $\hat{\mathbf{C}}_H \in \{0,1\}^{N_H \times 1}$.

\subsection{Multi-Level Conditioned Diffusion}
\label{subsec:MLCD}
The core of our method is a Transformer-based Denoising Diffusion Probabilistic Model (DDPM) \citep{ho2020denoising} that 
synthesizes hand pose parameters $\hat{\mathbf{H}}$ conditioned on the object point cloud $\mathbf{P}_O$, multi-level text $\mathbf{T}$, and predicted contact maps $\hat{\mathbf{C}}$. 
Instead of predicting noise, our model $f_\theta$ is trained to directly predict the denoised data $\hat{\mathbf{x}}_0 = f_\theta(\mathbf{x}_t, t, \mathbf{y})$, 
optimized with an L2 loss on the pose parameters: $L_{{diff}} = \mathbb{E}_{t,\epsilon}\big[\|\hat{\mathbf{x}}_0 - \mathbf{x}_0\|^2\big]$

\textbf{Condition Generation: Transformer Inputs.}
To achieve precise control, our model extracts multi-level conditional features from both geometric and textual modalities. 
On the geometric side, we use PointNet to extract global features $\mathbf{F}^{O}_{{glb}}, \mathbf{F}^{H}_{{glb}}$ 
and point-wise local features from the object point cloud $\mathbf{P}_O$, the initial hand point cloud $\mathbf{P}_H^0$, and the predicted contact maps $\hat{\mathbf{C}}$ from the previous stage. 
To focus on interaction regions, we leverage $\hat{\mathbf{C}}$ to adaptively select features of $N_{{loc}}^O=128$ object points and $N_{{loc}}^H=64$ hand points near contact areas, 
yielding $\tilde{\mathbf{F}}^{O}_{{loc}}$ and $\tilde{\mathbf{F}}^{H}_{{loc}}$. 
On the textual side, we utilize $f_{{text}}$ to extract both coarse-grained $\mathbf{F}_{{qwen}}^{{SSC}} = f_{text}(\mathbf{T}_{{SSC}})$ and fine-grained $\mathbf{F}_{{qwen}}^{{DSC}}$ text features.

\textbf{Conditional Injection: Coarse-to-Fine Control.}
We inject these features into the $N_{{inj}}=8$ blocks of our Transformer model in a hierarchical, coarse-to-fine fashion. This design ensures that global context, defined by SSCs and global geometry, shapes the overall pose in early denoising stages, while local details, defined by DSCs and contact-point features, are refined in later stages. Specifically, for the $i$-th Transformer block:\\
Early Stages ($i < 4$): Global context is injected, with no local features.
\begin{equation}\mathbf{y}_{{glb}}^i = {concat}(\mathbf{F}^{O}_{{glb}}, \mathbf{F}^{H}_{{glb}}, s_O, \mathbf{F}_{{qwen}}^{SSC}, t), \quad \mathbf{y}_{{loc}}^i = \emptyset \end{equation}
Later Stages ($4 \leq i < N_{{inj}}$): Local details are injected, switching to fine-grained conditions.
\begin{equation}
\mathbf{y}_{{glb}}^i = {concat}(\mathbf{F}^{H}_{{glb}}, s_O, \mathbf{F}_{{qwen}}^{DSC}, t), \quad \mathbf{y}_{{loc}}^i = {concat}(\tilde{\mathbf{F}}^{O}_{{loc}}, \tilde{\mathbf{F}}^{H}_{{loc}})
\end{equation}
To prevent over-reliance on any single condition and enhance robustness, we randomly drop each component of the global condition with a 10\% probability during training.
Within each block, the global condition $\mathbf{y}_{{glb}}^i$ modulates the main features via FiLM \citep{perez2018film}, 
while the local condition $\mathbf{y}_{{loc}}^i$ is integrated through cross-attention to provide fine-grained spatial cues. 
This dual mechanism effectively decouples global contextual guidance from local geometric refinement. 
Finally, the updated latent goes through self-attention and a Feed-Forward Network (FFN).

\textbf{Training Loss.}
To improve training stability and spatial alignment, we introduce two auxiliary losses alongside the primary diffusion loss $L_{{diff}}$. 
A global pose loss directly supervises the hand's global rotation $\mathbf{r}_{{rot}}$ and translation $\mathbf{T}$ 
to prevent overall pose drift, an issue exacerbated when directly regressing $\mathbf{H}$, which comprises parameters with disparate numerical ranges 
(e.g., shape $\boldsymbol{\beta}$, pose $\boldsymbol{\Theta}$, $\mathbf{r}_{{rot}}$, $\mathbf{T}$). 
A distance map loss ensures precise contact by supervising the distance map $\mathbf{d}_{{map}} \in \mathbb{R}^{21 \times N_O}$ 
from the 21 hand joints to the object surface. The final objective is a weighted sum:
\begin{equation}
    L_{{total}} = L_{{diff}} + \lambda_{{global}} \Big(| \hat{\mathbf{r}}_{{rot}} - \mathbf{r}_{{rot}}^{{gt}} | + | \hat{\mathbf{T}} - \mathbf{T}^{{gt}} |\Big) + \lambda_{{dmap}} | \hat{\mathbf{d}}_{{map}} - \mathbf{d}_{{map}}^{{gt}} |
\end{equation}

\subsection{Physical Constraints Refinement}
\label{subsec:PCR}
To address the common issue of global pose drift in free-form HOI generation, where the hand often fails to make contact with the object, 
we introduce an efficient physical refinement module. This module is powered by a refiner network, $f_{{refiner}}$, which inherits the Transformer architecture of our diffusion model. 
The process begins with a single forward pass to rapidly correct the global positioning of the initial pose $\hat{\mathbf{H}}_{{diff}}$, establishing primary physical contact. 
Subsequently, this corrected pose undergoes $N_{{tta}}$ iterations of test-time optimization (TTA) to fine-tune local contact details, such as finger placements.

The entire optimization is guided by our self-supervised cycle-consistency loss ($L_{cyc}$), 
which enforces bidirectional mapping consistency between hand and object contact surfaces. The core idea is that a hand contact point, 
after being mapped to the nearest object point via $\Phi$ (hand-to-object), should map back to its original location via the reverse mapping $\Psi$ (object-to-hand), and vice versa. 
This loss acts as a powerful regularizer, effectively reducing the ambiguity inherent in the mappings. 
We combine this with  $L_{{phy}}$ (see Eq.~\ref{eq:hand_refine}). The total refinement loss is defined as:
\begin{equation}
L_{{refiner}} =  L_{{phy}} + \lambda_{cyc}(\mathbb{E}_{\mathbf{P}_h \in \mathbf{P}_{C_H}} || \Psi(\Phi(\mathbf{P}_h)) - \mathbf{P}_h ||_1 +  \mathbb{E}_{\mathbf{P}_o \in \mathbf{P}_{C_O}} || \Phi(\Psi(\mathbf{P}_o)) - \mathbf{P}_o ||_1)
\end{equation}
\section{EXPERIMENTS}
\subsection{Experimental Settings}
Our experiments are conducted on the \dstname{} dataset. For each hand part contact category, we perform a random 4:1 split, yielding approximately 3.7k training and 677 test samples. 
To address the long-tailed distribution of hand part labels, we aggregate 10 less frequent hand part categories 
and then apply resampling using unique 7-bit labels to balance the data.
The model is trained for 1000 epochs using the Adam optimizer with a learning rate of 1e-4 and a batch size of 128. 
The diffusion model's parameters are frozen during the training of the refiner module. We evaluate our method from four perspectives: 
(1) Contact Accuracy, assessed by IoU(P-IoU) and F1-score(P-F1) against ground-truth contacts parts. 
(2) Physical Plausibility, measured by Mean Per-Vertex Position Error (MPVPE), Penetration Depth (PD), and Penetration Volume (PV). 
Note that unlike works focusing on grasping \citep{jiang2021graspTTA}, we do not employ physics engine-based stability simulation metrics, 
as our scope of interactions is broader than force-closure grasps. (3) Diversity, quantified by entropy and cluster size. 
(4) Semantic Consistency, evaluated using a point cloud-based FID (P-FID) \citep{nichol2022pointegenerating3dpoint}, VLM assisted evaluation, and a perceptual score (PS) from 10 users.

\begin{figure*}[t]
	\centering
	\footnotesize
    \renewcommand{\arraystretch}{1.}
        \begin{overpic}
            [width=1\linewidth]{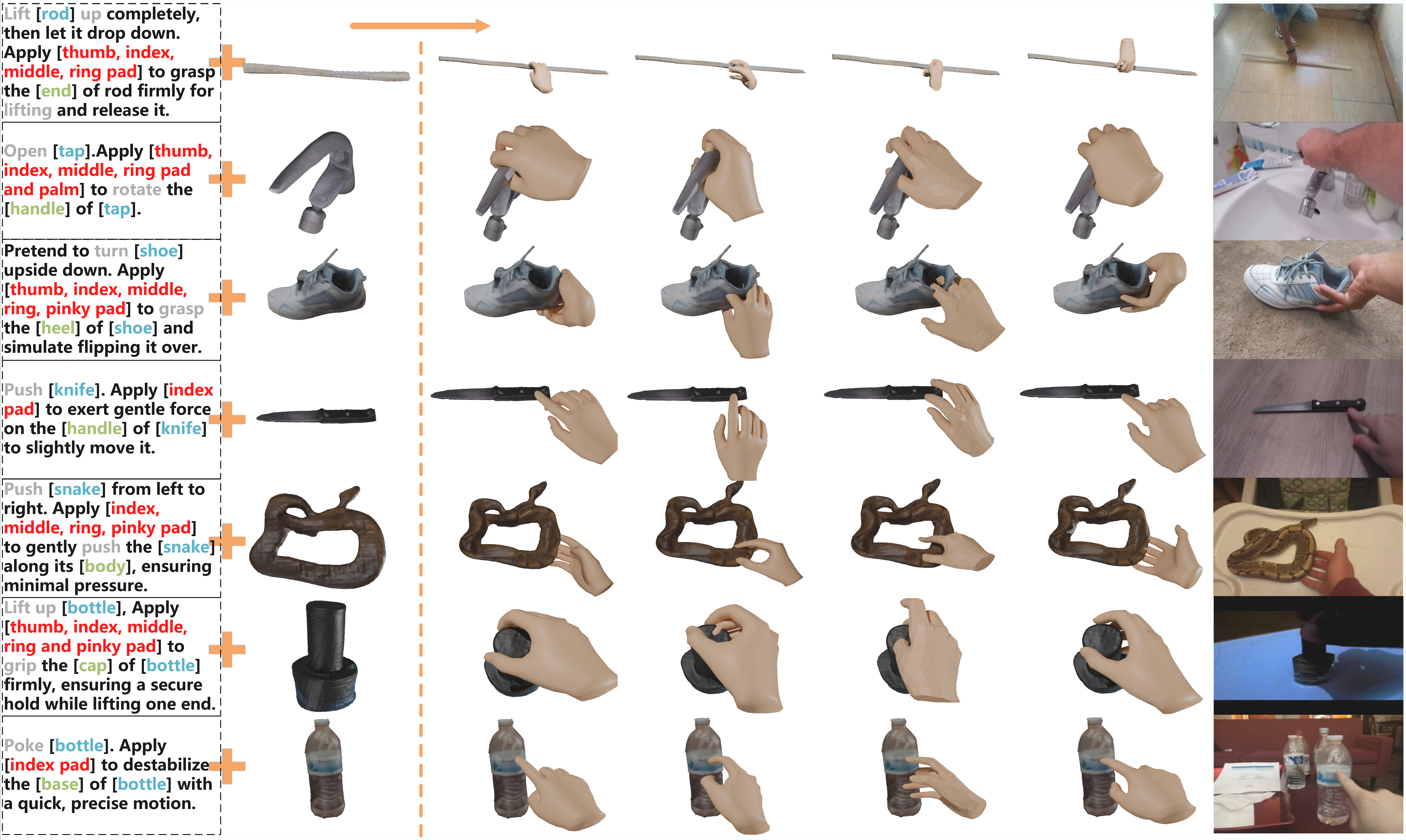}
        \put(5,61){\textbf{\texttt{SSCs}}}
        \put(20,61){\textbf{\texttt{Obj}}}
        \put(35,61){\textbf{\texttt{Ours}}}
        \put(45,61){\textcolor[rgb]{0.99,0.5,0.0}{\textbf{\texttt{ContactGen}}}}
        \put(60,61){\textcolor[rgb]{0.2,0.8,0.1}{\textbf{\texttt{Text2HOI}}}}
        \put(77,61){\textbf{\texttt{GT}}}
        \put(90,61){\textbf{\texttt{$\mathbf{I}_{hoi}$}}}
        \end{overpic}
	\caption{\textbf{Visualization Results.} 
    Comparisons of different methods on samples from the \dstname{} test set. 
    Each sample consists of SSCs and an object mesh as input, with the output being an interactive hand pose.
    The last row shows the original authentic 2D HOI frame from internet videos.
    }
 \label{fig:main_results}
\vspace{-10pt}
\end{figure*}
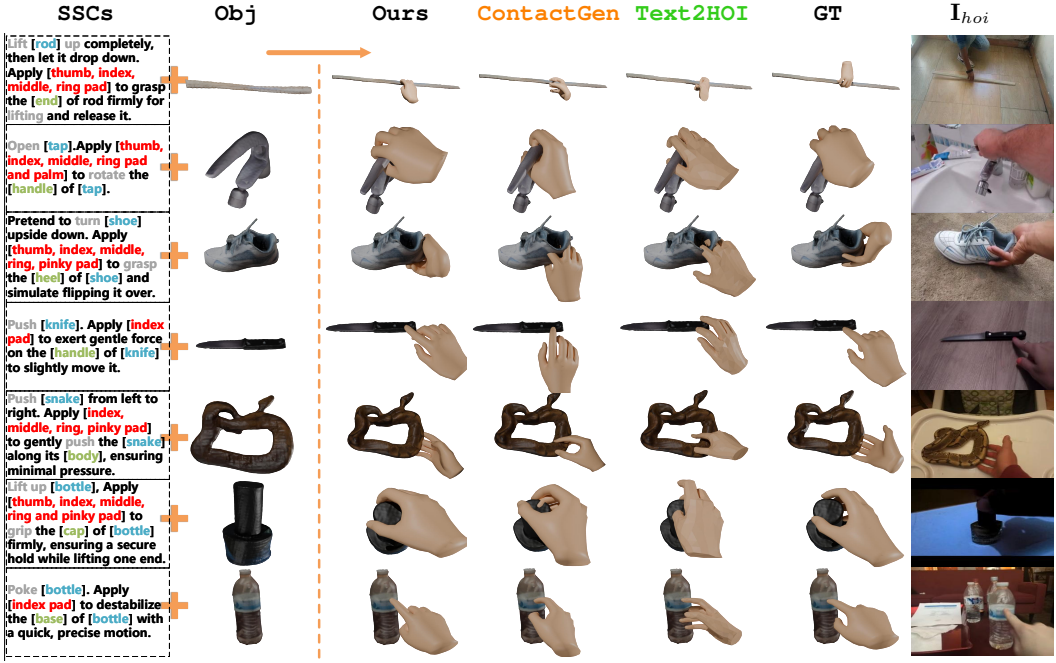

\subsection{Comparisons}
As existing methods have not explored fine-grained controlled HOI generation, we select two representative types of baselines:
(1) \textit{ContactGen} \citep{contactgen}: an object-conditioned multi-layer CVAE using coarse hand part labels.
(2) \textit{Text2HOI} \citep{cha2024text2hoi}: a transformer-based conditional diffusion model guided by coarse text conditions. We remove its temporal axis and adapt it for our setting.
Compared to typical grasping datasets, hand poses in \dstname{} exhibit higher degrees of freedom. Both baseline methods exhibit noticeable overall hand drift.
To ensure fair comparison, we also augment them with an optimization-based post-processing module to correct hand poses.
Experimental results in Table~\ref{tab:cmp_exp_main} show that our method outperforms baselines across most metrics.
Visual results in Fig.~\ref{fig:main_results} further demonstrate that our method generates more realistic HOI poses that better align with input text descriptions.
\renewcommand{\arraystretch}{1.2}
\begin{table*}[h]
  \centering
  \setlength{\extrarowheight}{1pt} 
  \setlength{\tabcolsep}{4pt} 
  \small
  \begin{tabular}{@{}l|cc|ccc|cc|ccc@{}}
    \toprule    
    {\textbf{Method}} & \multicolumn{2}{c|}{\textbf{Contact Acc.}} & \multicolumn{3}{c|}{\textbf{Physical Plausibility}} & \multicolumn{2}{c|}{\textbf{Diversity}} & \multicolumn{3}{c}{\textbf{Semantic Consistency}} \\
    \cmidrule(l){2-3} \cmidrule(l){4-6} \cmidrule(l){7-8} \cmidrule(l){9-11}
    & P-IoU$\uparrow$ & P-F1$\uparrow$ & MPVPE$\downarrow$  & PD$\downarrow$ & PV$\downarrow$ & Ent.$\uparrow$ & CS$\uparrow$ & P-FID$\downarrow$ &VLM$\uparrow$ & PS$\uparrow$ \\
    \midrule
    ContactGen        &0.620&0.730&5.46&1.296&7.37&2.85&4.93&6.08 &4.8 &6.3 \\ 
    Text2HOI          &0.711&0.795&4.69&1.239&4.93&2.85&5.20&15.72&6.5 & 7.5 \\ 
    \textbf{Ours} \cellcolor{mygray}   &\cellcolor{mygray}0.776&\cellcolor{mygray} 0.844&\cellcolor{mygray}2.97&\cellcolor{mygray}0.932&\cellcolor{mygray}2.67&\cellcolor{mygray} 2.93&\cellcolor{mygray} 5.40&\cellcolor{mygray} 4.13 &\cellcolor{mygray}7.1&\cellcolor{mygray}8.8 \\ % 
    \bottomrule
  \end{tabular}
  \caption{\textbf{Quantitative comparison on hand-object interaction synthesis.} We evaluate all methods on our test set using comprehensive metrics covering physical plausibility, contact accuracy, diversity, and semantic consistency. $\uparrow$ indicates higher is better, $\downarrow$ indicates lower is better.}
  \label{tab:cmp_exp_main}
    \vspace{-20pt} 
\end{table*}

\subsection{Ablation Study}
\begin{figure}[h!]
  \centering
  \footnotesize
  \begin{minipage}[c]{0.68\linewidth}
    \centering
    {    
      \renewcommand{\arraystretch}{1.}
      \begin{tabular}{c|ccccccc}
      \toprule
      \textbf{Metrics}             &P-IoU$\uparrow$&P-F1$\uparrow$ &MPVPE$\downarrow$& PD$\downarrow$ & PV$\downarrow$ &P-FID$\downarrow$ \\ \midrule
      \textbf{Ours(\xmark{} TTA)}    &0.728\cellcolor{mygray} &0.805\cellcolor{mygray} &3.00 &1.093 &4.82 &4.84 \cellcolor{mygray} \\ \midrule
      \xmark{} hoc.                  &0.492 &0.611 &4.93 &1.330 &5.50 &5.41  \\
      \xmark{} refiner               &0.513 &0.621 &5.05 &0.723\cellcolor{mygray} &2.98\cellcolor{mygray} &5.84  \\ 
      \xmark{} $L_{cyc}$.            &0.702 &0.787 &3.00 &1.100 &5.29 &5.79  \\ \midrule
      \xmark{} mul.                  &0.525 &0.631 &5.00 &1.464 &6.52 &6.84  \\ 
      \xmark{} $T_{DSC}$             &0.698 &0.784 &3.02 &1.119 &5.28 &6.09  \\
      \xmark{} $T_{SSC}$            &0.687 &0.778 &2.92 &1.119 &5.17 &5.52  \\\hline\hline
      CLIP   & 0.713 & 0.798 & 2.87\cellcolor{mygray} & 1.136 & 4.85 & 4.84 \\
      BERT   & 0.705 & 0.790 & 2.91 & 1.182 & 4.99 & 6.08 \\
      MPNet  & 0.704 & 0.788 & 2.87\cellcolor{mygray} & 1.114 & 5.06 & 6.02 \\      \bottomrule    
      \end{tabular}
    }
    \captionof{table}{Analyzes the contributions of various components, including the absence of $\mathbf{M}_O$ and $\mathbf{M}_H$ (hoc.) for guiding spatial relationship generation, the multi-level network structure (mul.), and the multi-level text. }
    \label{subtab:ablation}
  \end{minipage}%
  \hfill
  \begin{minipage}[c]{0.28\linewidth}
    \centering
    % \vspace{1pt}
    \includegraphics[width=\linewidth]{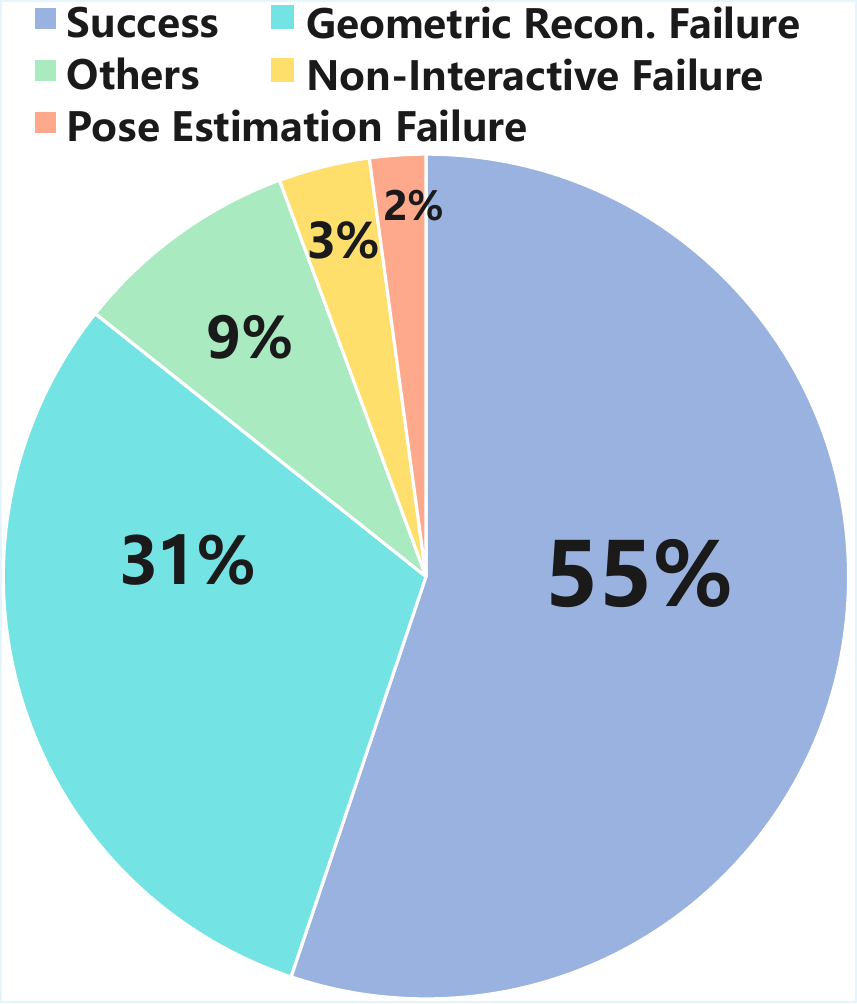}
    \caption{Breakdown of \dstname{} reconstruction outcomes.}
    \label{subfig:success_rate}
  \end{minipage}
  \vspace{-10pt} 
\end{figure}

\begin{figure}[htbp]
  \centering
  \begin{minipage}[c]{0.4\linewidth}
    \centering
    \includegraphics[width=\linewidth]{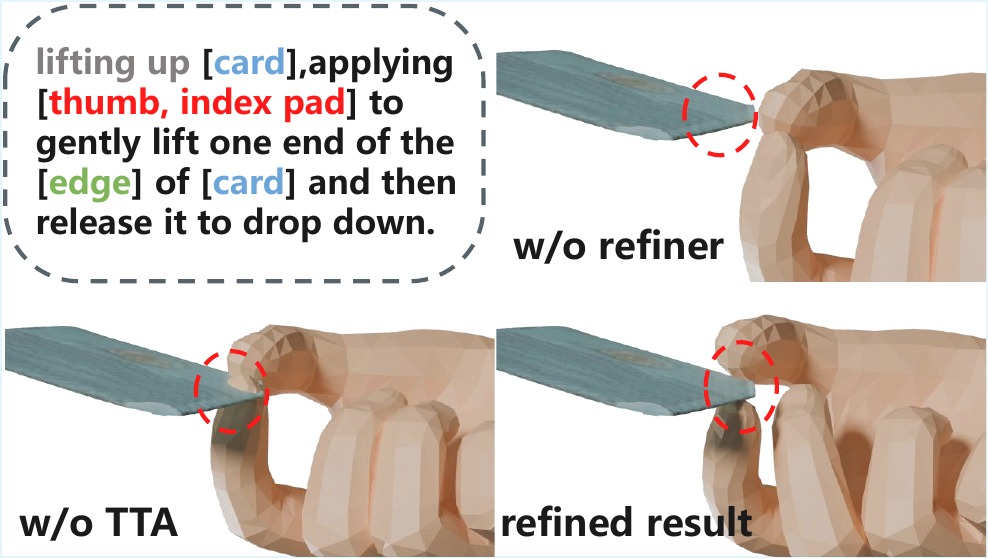}
    \caption{Contact guidance visualization.}
    \label{fig:contact_ablation}
  \end{minipage}%
  \hfill
  \begin{minipage}[c]{0.58\linewidth}
    \centering
    \includegraphics[width=\linewidth]{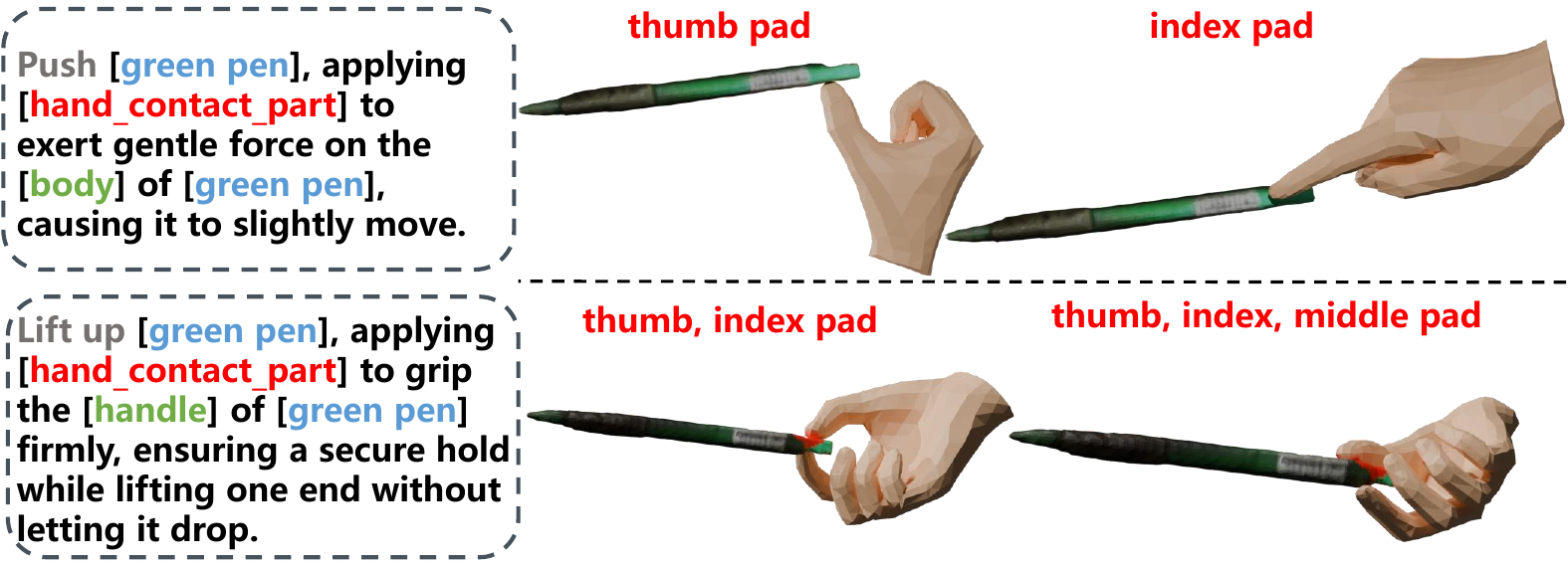}
    \caption{Generation results for a "ballpoint pen." Each subfigure shows a unique and plausible hand pose generated by specifying different control signals.}
    \label{fig:action_analysis}
  \end{minipage}
  \vspace{-10pt} 
\end{figure}

To evaluate the effectiveness of our contact-guided generation of spatial relations in HOI and the coarse-to-fine text control design, 
we conduct ablation studies as shown in Table~\ref{subtab:ablation}, \xmark{} means without. For clarity, TTA is disabled. 
We argue for the primacy of contact metrics, as penetration metrics, PD and PV are meaningful only after hand-object contact is established; 
otherwise, they can be misleading. This is starkly exemplified by the "\xmark{} refiner" variant, 
which scores poorly on contact yet achieves deceptively low PD/PV values simply because the generated hand drifts away from the object, thus avoiding interaction entirely.
This distinguishes our complex task from traditional grasping, where contact is facilitated by priors. 
The consistent degradation in contact performance upon removing any module confirms their synergistic importance. Fig.~\ref{fig:contact_ablation} offers a qualitative visualization of the step-by-step improvements afforded by our contact guidance.

\subsection{Discussion}
\textbf{Dataset Reconstruction Analysis. }
Our dataset reconstruction from in-the-wild images achieved an approximate success rate of 55\%. The failures, detailed in Fig.~\ref{subfig:success_rate}, 
provide insight into the inherent challenges of this task. The primary obstacle is geometric reconstruction failure, where 3D mesh recovery is unsuccessful. 
Further examples are available in the Appendix\ref{sec:recon_analysis}.

\textbf{Text-guided Model and Usage.}
We replace the text encoder with other common token- or sentence-level encoders (e.g., CLIP \citep{clip}, BERT \citep{devlin2019bert}, MPNet \citep{song2020mpnet}) to analyze their impact on generation quality. 
Results indicate that Qwen-7B offers better performance in capturing fine-grained semantic details, detailed in Tab.~\ref{subtab:ablation}.

\textbf{Generation Under Different Control Conditions for the Same Object.}
Using the same object under varying hand contact regions and interaction intents, our model produces diverse and physically plausible hand poses, as visualized in Fig.~\ref{fig:action_analysis}.

\section{CONCLUSION}
In this paper, we addressed the limitations of grasp-centric approaches by introducing the Free-form HOI Generation task. Our work expands the synthesis paradigm beyond simple grasping to a broader, more semantically expressive spectrum of interactions. To support this, we built an automated pipeline to construct \dstname{}, an in-the-wild 3D dataset for daily HOIs, providing a critical resource to enable future research in this domain.

\textbf{Limitations and Future Directions.} Our framework currently focuses on static HOI snapshots, which inherently limits its ability to capture the temporal dynamics of an interaction process. While our pipeline offers rapid expansion, the current dataset scale also presents an area for future growth. In the future, we plan to extend our work to dynamic sequences by leveraging large-scale video datasets and incorporating 6-DoF object pose estimation, thus modeling the entire human-environment interaction process.

\bibliography{iclr2026_conference}
\bibliographystyle{iclr2026_conference}
\appendix
\newpage
\section{Appendix}
\subsection{More Experiment Results}
\begin{figure}[t]
	\centering
	\includegraphics[width=0.9\textwidth]{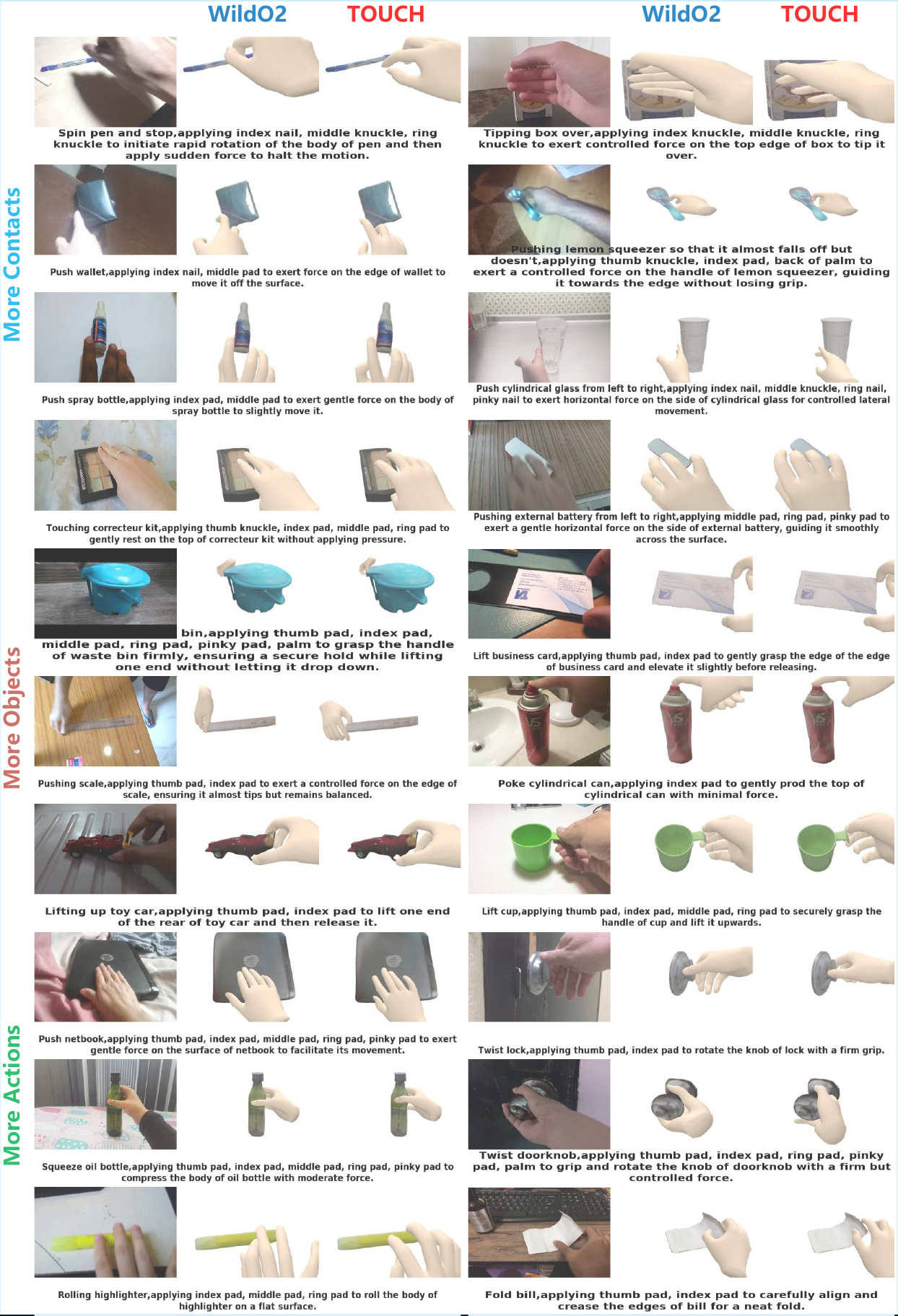}
\caption{\textbf{Visualization Results of more \dstname{} test set samples.} For each sample, the leftmost image represents the hand-object interaction frame ($\mathbf{I}_{hoi}$) extracted from Internet videos. The middle image shows our reconstruction of $\mathbf{I}_{hoi}$, which serves as the ground truth for WildO2. The rightmost image illustrates the free-form HOI generated by our method, TOUCH, conditioned on object meshes and DSCs. The bottom row contains the corresponding DSCs.}
 \label{fig:more_results}
\vspace{-10pt}
\end{figure}

We performed a per-category statistical analysis of the metrics for the most frequent verb categories in the test set, as detailed in Table \ref{tab:verb_stats}. 
The results reveal significant variations across different verb categories, reflecting the distinct characteristics of each action.

For instance, the Poke category, which typically involves a singular or highly localized contact area, 
consequently yields higher scores on contact-related metrics such as P-IoU and P-F1. 
However, this action imposes fewer constraints on the overall hand orientation and the configuration of non-contacting fingers, 
leading to a higher MPVPE for the generated hand poses. 
Conversely, the Lift category, which closely resembles a simple grasping motion, is generally less complex. 
As a result, it demonstrates strong performance across most metrics, which corroborates our earlier assertion regarding the relative simplicity of grasping operations. 
In general, for a given verb, greater diversity in the associated actions correlates with increased learning difficulty for the model.

\begin{table}[h]
\centering
\renewcommand{\arraystretch}{1.2}
\setlength{\tabcolsep}{4pt}
\begin{tabular}{l|ccccccccccc}
\toprule
 & Push & Poke & Lift & Pick & Move & Turn & Pull & Put & Open & Roll & Spin \\
\midrule
P-IoU & 0.762 & 0.811 & 0.805 & 0.752 & 0.773 & 0.725 & 0.830 & 0.753 & 0.778 & 0.739 & 0.763 \\
P-F1 & 0.833 & 0.852 & 0.872 & 0.834 & 0.844 & 0.821 & 0.891 & 0.828 & 0.866 & 0.819 & 0.834 \\
MPVPE & 3.098 & 3.458 & 2.724 & 2.511 & 2.618 & 3.014 & 2.301 & 3.034 & 2.621 & 2.653 & 3.844 \\
PD & 1.135 & 0.391 & 0.568 & 1.127 & 0.811 & 1.462 & 1.214 & 1.105 & 1.206 & 1.294 & 0.892 \\
PV & 3.025 & 0.922 & 1.716 & 2.814 & 2.842 & 5.718 & 3.440 & 3.656 & 4.081 & 1.652 & 3.052 \\
\bottomrule
\end{tabular}
\caption{Per-category performance metrics for the most frequent verb categories in the test set.}
\label{tab:verb_stats}
\end{table}

Additionally, we evaluate our method on the OakInk dataset. Using its CapGrasp~ \citep{li2024semgrasp} annotations and computed part contact information, 
we construct fine-grained text conditions through summarization via Qwen~ \citep{qwen} and verification.

\renewcommand{\arraystretch}{1.2}
\begin{table*}[h]
  \centering
  \setlength{\extrarowheight}{1pt} 
  \setlength{\tabcolsep}{4pt}
  \small
  \begin{tabular}{@{}l|cc|ccc|cc|c@{}}
    \toprule    
    {\textbf{Method}} & \multicolumn{2}{c|}{\textbf{Contact Acc.}} & \multicolumn{3}{c|}{\textbf{Physical Plausibility}} & \multicolumn{2}{c|}{\textbf{Diversity}} & \multicolumn{1}{c}{\textbf{Semantic}} \\
     \cmidrule(l){2-3} \cmidrule(l){4-6} \cmidrule(l){7-8} \cmidrule(l){9-9} 
    & P-IoU$\uparrow$ & P-F1$\uparrow$ & MPVPE$\downarrow$  & PD$\downarrow$ & PV$\downarrow$ & Ent.$\uparrow$ & CS$\uparrow$ & P-FID$\downarrow$ \\
    \midrule
    ContactGen        &0.654&0.769&6.69&\cellcolor{mygray} 0.790&16.59&\cellcolor{mygray} 2.93&\cellcolor{mygray} 5.33&11.25 \\ 
    Text2HOI          &0.778&0.860&5.49&0.807&7.05&2.84&3.48&4.95 \\ 
    \textbf{Ours}\cellcolor{mygray} &\cellcolor{mygray}0.812&\cellcolor{mygray}0.882&\cellcolor{mygray} 4.49&0.939&\cellcolor{mygray}5.89&2.92&3.50&\cellcolor{mygray} 3.21 \\ 
    \bottomrule
  \end{tabular}
  \caption{\textbf{Quantitative comparison on hand-object interaction synthesis in OakInk-Shape dataset.}  $\uparrow$ indicates higher is better, $\downarrow$ indicates lower is better.}
  \label{tab:cmp_exp}
    % \vspace{-20pt} 
\end{table*}

\subsection{Details of \dstname{} acquisition}
\label{sec:dataset_details}

\subsubsection{Reconstructable Clip Selection from Wild Videos}
Our data generation process commences with the Something-Something V2 (SSv2) dataset  \citep{sthv2}, a large-scale collection of over 220,847 videos. 
We chose this dataset for several key reasons. First, its vast scale and diversity, resulting from a crowdsourced effort where contributors enact specific verb-noun prompts, 
provide a rich variety of goal-oriented hand-object interactions. Second, many videos feature close-up shots of the hand-held object, which is advantageous for 3D reconstruction. 
Critically, the associated Something-Else dataset  \citep{sthelse} furnishes bounding box annotations for hands and objects in approximately 180,049 of these clips, 
offering an invaluable starting point. However, a primary challenge of SSv2 is its low resolution (typically 240 pixels in height), 
which necessitates a rigorous filtering pipeline to isolate clips suitable for high-fidelity reconstruction.

To ensure the viability of our reconstruction pipeline, we implemented a multi-stage filtering process to distill the initial 180k annotated clips into a high-quality subset. 
The screening criteria were designed to simplify the interaction scenario (e.g., single hand, single object) and guarantee sufficient visual quality for reliable 3D modeling. 
This process effectively removes clips that are ambiguous, occluded, or otherwise intractable. The detailed steps and their impact on the dataset size are summarized in Table~\ref{tab:filtering_pipeline}.

\begin{table}[h!]
\centering
\caption{The multi-stage filtering pipeline for selecting reconstructable clips from the Something-Something V2 dataset. The process starts with clips annotated by Something-Else and progressively refines the selection based on interaction complexity and visual quality.}
\label{tab:filtering_pipeline}
\resizebox{\textwidth}{!}{%
\begin{tabular}{@{}lllc@{}}
\toprule
\textbf{Step} & \textbf{Criterion} & \textbf{Description} & \textbf{Clips Remaining} \\
\midrule
1 & Initial Annotated Set & Clips from SSv2 with hand and object bounding box annotations. & 180,049 \\
2 & Single-Hand Interaction & Exclude clips involving multi hands to simplify interaction modeling. & 137,578 \\
3 & Single-Object Interaction & Retain only clips where a single object is being manipulated. & 82,728 \\
4 & Minimum Object Size & Discard clips where the object is too small to be reliably reconstructed. & 57,384 \\
5 & Object Visibility & Exclude clips where the object is too large or moves out of frame. & 12,888 \\
6 & Frame Stability & Ensure stable pre-interaction and interaction frames can be identified. & 8,551 \\
\bottomrule
\end{tabular}%
}
\end{table}

\subsubsection{Acquisition of O2HOI Frame Pairs}

This section details the automated procedure for extracting an "Object-only to Hand-Object Interaction" (O2HOI) frame pair, denoted as $(I_{{ref}}, I_{{hoi}})$, from each video clip.

\textbf{1. Mask Generation and Frame Selection Preliminaries.}
For each frame in a given video sequence, we generate object masks ($M_O$) and hand masks ($M_H$) using the SAM2.
The interaction period, $\mathcal{T}_{{HOI}} = [i_{{first}}, i_{{last}}]$, is defined as the contiguous block of frames where the Intersection over Union (IoU) between the object mask and the hand mask, each expanded by several pixels, exceeds a predefined threshold.
The reference frame, $I_{{ref}}$, is selected as the object-only frame temporally closest to this interaction period. \\

\textbf{2. Optimal HOI Frame Selection. }
The primary objective is to select an index $i_{{hoi}}$ from $\mathcal{T}_{{HOI}}$ such that 
the object's pose in frame $I_{{hoi}}$ has undergone minimal change relative to its pose in $I_{{ref}}$. 
We achieve this by estimating the 2D affine transformation between the object in the reference frame and in every candidate frame within $\mathcal{T}_{{HOI}}$. 
This estimation utilizes the RoMa dense feature matcher  \citep{edstedt2024roma}. 
The detailed steps for comparing a candidate frame $I_t$ (where $t \in \mathcal{T}_{{HOI}}$) to the reference frame $I_{{ref}}$ are as follows:
\begin{itemize}
    \item Feature Matching: For each pair $(I_{{ref}}, I_t)$, we extract robust keypoint correspondences $(\mathbf{k}_{{ref}}, \mathbf{k}_t)$ within their respective object masks, $M_{{ref}}^O$ and $M_t^O$.
    \item Transformation Estimation: The keypoints are used to robustly estimate an affine transformation matrix $\mathbf{A}_t$ using RANSAC.
    \item  Rotation and IoU Calculation: The rotation angle sequence $\theta_t$ and IoU metrics are computed as follows:
    \begin{align}
        R_S = \mathbf{A}_t[0:2, 0:2], \quad U, S, V^\top &= {SVD}(R_S), \quad R = UV^\top \\
        \theta_t = \arctan2(R_{21}, R_{11}) \cdot \frac{180}{\pi}, &\quad {IoU}_t = \frac{|{warpAffine}(M_{{ref}}^O, \mathbf{A}_t) \cap M_t|}{|{warpAffine}(M_{{ref}}^O, \mathbf{A}_t) \cup M_t|}
    \end{align}
    \item Interaction Frame Selection: with IoU gradient $\Delta {IoU}_t$. The selected HOI frame index $i_{{hoi}}$ is determined by:
        \begin{equation}
        i_{\text{hoi}} = 
            \begin{cases}
            \arg\min_{t \in \mathcal{T}_{\text{HOI}}} |\theta_t|, & \text{if } \min_{t \in \mathcal{T}_{\text{HOI}}} |\theta_t| > \text{MAX\_MIN\_ANGLE} \\
            \arg\min_{t \in \mathcal{S}_{\text{stable}}} |t - i_{\text{max}}|, & \text{if } \max_{t \in \mathcal{T}_{\text{HOI}}} |\theta_t| < \text{MIN\_MAX\_ANGLE} \\
            \arg\min_{\substack{t \in \mathcal{T}_{\text{HOI}} \\ |\theta_t| < \text{MAX\_MIN\_ANGLE}}} |t - i_{\text{max}}|, & \text{otherwise}
            \end{cases}
        \end{equation}
        where $\mathcal{S}_{\text{stable}} = \{ t \in \mathcal{T}_{\text{HOI}} \mid |\Delta \mathrm{IoU}_t| < \text{DT\_IOU\_THRES} \}$, and MAX\_MIN\_ANGLE=1, MIN\_MAX\_ANGLE=5
    \item Inpainting Mask Generation: In summary, we obtain the O2HOI frame pair $(i_{{ref}}, i_{{hoi}})$ and the corresponding inpainting mask for subsequent processing:
    \begin{equation}M_{{inpaint}} = {warpAffine}(M_{{ref}}^O, \mathbf{A}_{{hoi}}) \end{equation}
\end{itemize}
This procedure ensures accurate spatial alignment between the object-only and interaction frames, facilitating efficient and reliable inpainting mask generation for downstream tasks.

\subsubsection{Estimation of Object Elevations}
The backbone of InstantMesh \citep{instantmesh}, Zero123 \citep{zero123}, generates novel views of objects; however, the predicted elevation angles may not precisely align with the control signals. 
To address this, we employ a rendering-based matching strategy to estimate the optimal elevation angle for the reconstructed 3D object mesh. 
Specifically, we render the mesh from candidate elevations and select the angle whose 2D projection best matches the HOI frame, 
providing a reliable initialization for subsequent optimization. The search is performed in two stages: (a) a coarse search over the range $[-40^\circ, 60^\circ]$ with 11 sampled angles, 
and (b) a fine search within $[{best}-10^\circ, {best}+10^\circ]$ using 21 samples. The selection criterion is the point-wise mean squared error (MSE) between the rendered projection and the aligned HOI frame.
\begin{equation}
{elevation} = \arg\min_{\theta \in \Theta} \frac{1}{N} \sum_{i=1}^{N} (I_{{ref}}[i] - I_{{rendered}}(\theta)[i])^2
\end{equation}
We employ the CLIP-similarity metric to automatically assess the reconstruction quality of image-to-3D methods.
This enables efficient and coarse filtering of clips, ensuring that only those with high-quality and reliable 3D reconstructions 
are selected for subsequent processing and annotation.

\subsubsection{Reconstruction Analysis}
\label{sec:recon_analysis}
\begin{figure}[h]
\centering
\includegraphics[width=\textwidth]{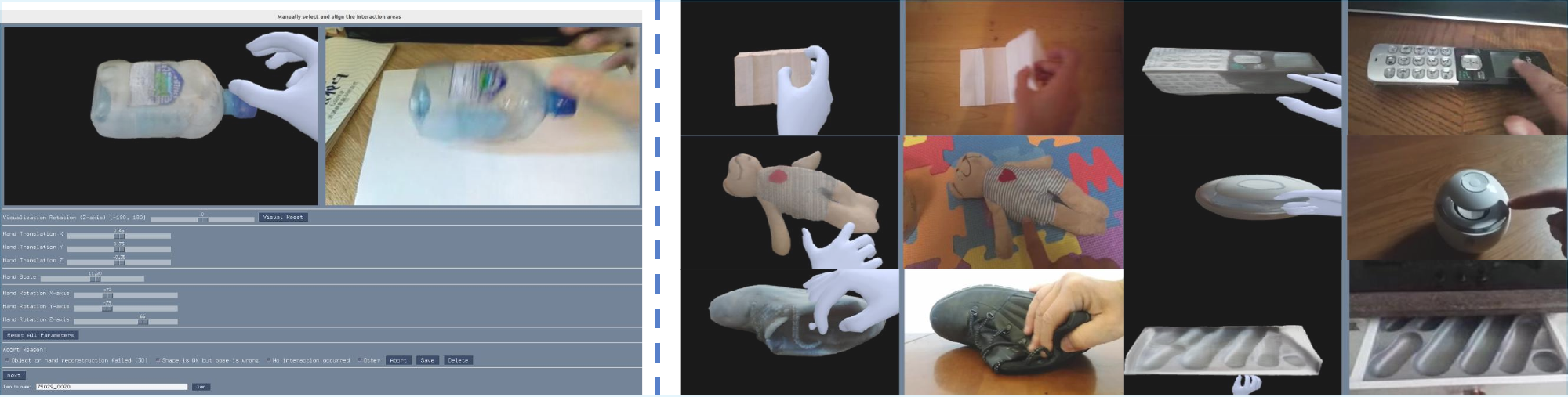}
\caption{
\textbf{Examples of reconstruction failures.}
(Left) A screenshot of our annotation UI showing a "Non-Interactive" failure, where the reconstructed hand and object do not interact.
(Right Top) Examples of "Geometric Reconstruction Failure" for the object and the hand. This was the most common failure category, with object geometry being more challenging to reconstruct than the hand's.
(Right Middle) Examples of "Pose Estimation Failure," where the 6DoF pose of the hand or object was incorrectly recovered.
(Right Bottom) Examples of "Other Cases." The first example shows a deformable object, and the second shows an object that is part of a larger structure (a drawer).
}
\label{fig:recon_fail}
\end{figure}

The primary obstacle is Geometric Reconstruction Failure, where 3D mesh recovery is unsuccessful. 
Another category, Non-Interactive Cases, is defined for instances where the 2D HOI guide image itself showed no discernible contact. 
Explicit failures in Pose Estimation account for 2.1\%. Finally, the Other Cases category contains various failures that bypassed initial rule-based screening; 
these include some of the most complex scenarios, such as interactions involving deformable or transparent objects, 
which fall outside the scope of our current reconstruction method. Details and examples of each failure type are provided in Figure~\ref{fig:recon_fail}.

\subsubsection{Contact Map Computation}

Given an object point cloud $\mathbf{O} \in \mathbb{R}^{N_o \times 3}$ and a hand point cloud $\mathbf{H} \in \mathbb{R}^{N_h \times 3}$, we propose a robust four-stage contact map computation algorithm:

\textbf{Stage 1: Bidirectional Nearest Neighbor Voting.} For each point, we compute the nearest neighbor in the opposite set and accumulate votes:
\[
{vote}_o(i) = \sum_{j} \mathbf{1}[{NN}(h_j) = o_i], \quad
{vote}_h(j) = \sum_{i} \mathbf{1}[{NN}(o_i) = h_j]
\]

\textbf{Stage 2: Candidate Selection.} High-frequency candidate points are selected based on the upper quantile $\alpha$ of the vote distribution:
\[
\mathcal{C}_o = \left\{ i \mid {vote}_o(i) \geq Q_{1-\alpha}({vote}_o) \right\}
\]

\textbf{Stage 3: Distance Validation.} Core contact points are further filtered by requiring their hand-object distance $d_i$ to be below both a quantile threshold $\beta$ and an absolute threshold $\epsilon$:
\[
\mathcal{S}_o = \left\{ i \in \mathcal{C}_o \mid d_i < \min(Q_\beta(d), \epsilon) \right\}
\]

\textbf{Stage 4: Region Expansion.} The contact region is expanded by including points within a radius $\gamma \cdot \bar{d}$ of any core contact point, where $\bar{d}$ is the mean $k$-nearest neighbor distance:
\[
\mathcal{M}_o = \left\{ i \mid \min_{s \in \mathcal{S}_o} \| o_i - o_s \|_2 \leq \gamma \cdot \bar{d} \right\}
\]

% We empirically set the parameters as follows: $\alpha=0.4$, $\beta=0.2$, $\epsilon=2 \times 10^{-5}$, and $\gamma=0.2$.所有点都不在距离阈值内时，按10\%分位取mask

\subsubsection{\dstname{} DSCs Prompt}
{\fontsize{8}{10}\selectfont
\begin{lstlisting}
[Structured Output Protocol]
As a hand-object interaction analyzer, generate JSON strictly like this:
{
    "obj_category": "cylindrical",
    "general": "grip [obj_category]",
    "physical": "Apply [hand_contact] to establish stable three-point contact with [obj_contact] of [obj_category] while other fingers form loose sphere.",
    "hand_contact": ["thumb pad", "pinky nail", "palm"],
    "obj_contact": "body"
}

[Key Requirements]
1. Mandatory placeholders: [hand_contact],[obj_contact],[obj_category]
2. Physical Field Rules:
    - Use ONLY [hand_contact] placeholder - ABSOLUTELY NO explicit contact point enumeration
    - must describe contact areas, interaction methods, and force levels considering object functionality
3. hand_contact is where interaction is most likely to occur.Input hand_contact may contain sensing errors from point cloud analysis. Valid hand_contact options (17 total):['thumb pad','index pad','middle pad','ring pad','pinky pad','thumb nail','index nail','middle nail','ring nail','pinky nail','thumb knuckle','index knuckle','middle knuckle','ring knuckle','pinky knuckle','palm','back of palm',]
4. ABSOLUTELY NO PREAMBLE/FOOTNOTES, only RAW JSON output

Current Input Interaction Description: "hand_contact": xxx, "general_intention": xxx.
\end{lstlisting}
}

\subsubsection{Dataset Analysis}
\begin{figure}[h]
    \centering
    \begin{minipage}{0.5\textwidth} 
        \centering
        \includegraphics[width=\linewidth]{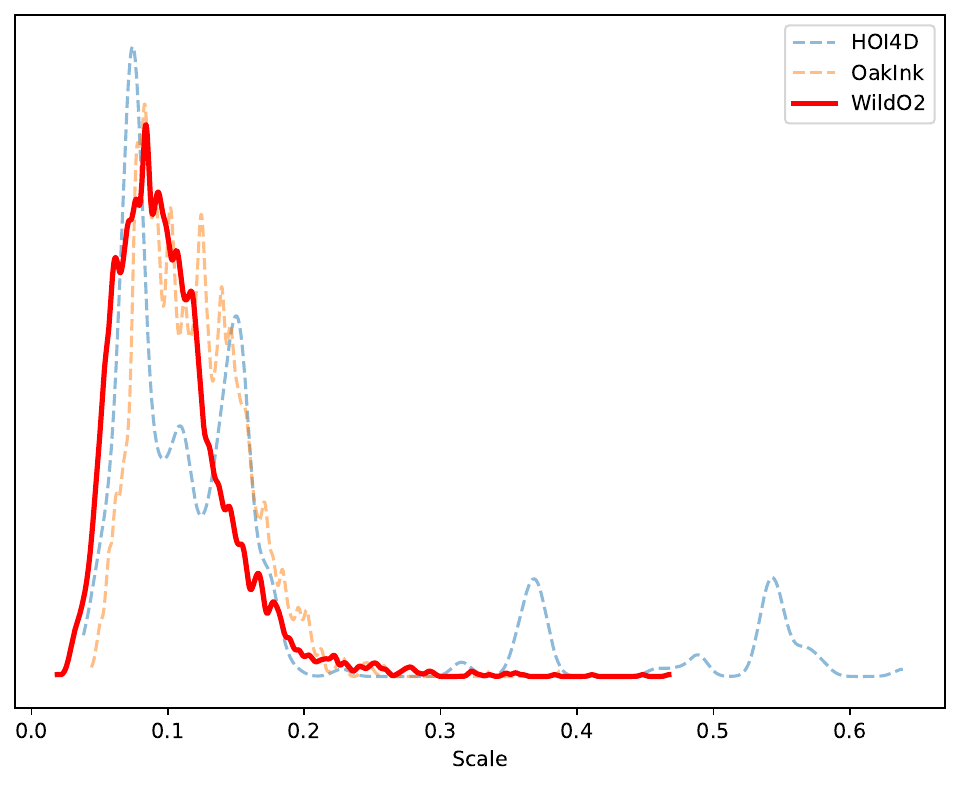}
    \end{minipage}\hfill
    \begin{minipage}{0.5\textwidth} 
        \centering
        \includegraphics[width=\linewidth]{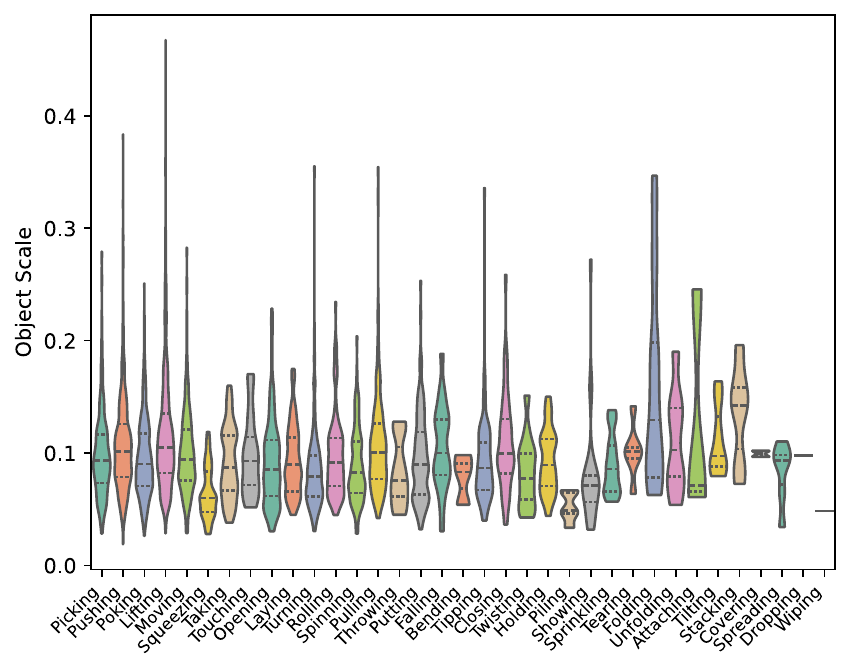}
    \end{minipage}
    \caption{The left figure is KDE curves of object scales in different datasets. Our dataset, collected without object category bias, shows a more uniform scale distribution., 
    while the right figure depicts the object size distribution across different verb categories.
    Actions like "squeezing" typically involve smaller objects, whereas "folding" tends to be associated with larger ones.}
  \label{fig:scale_dist}
\end{figure}

\begin{figure}[h]
  \centering
  \includegraphics[width=0.8\textwidth]{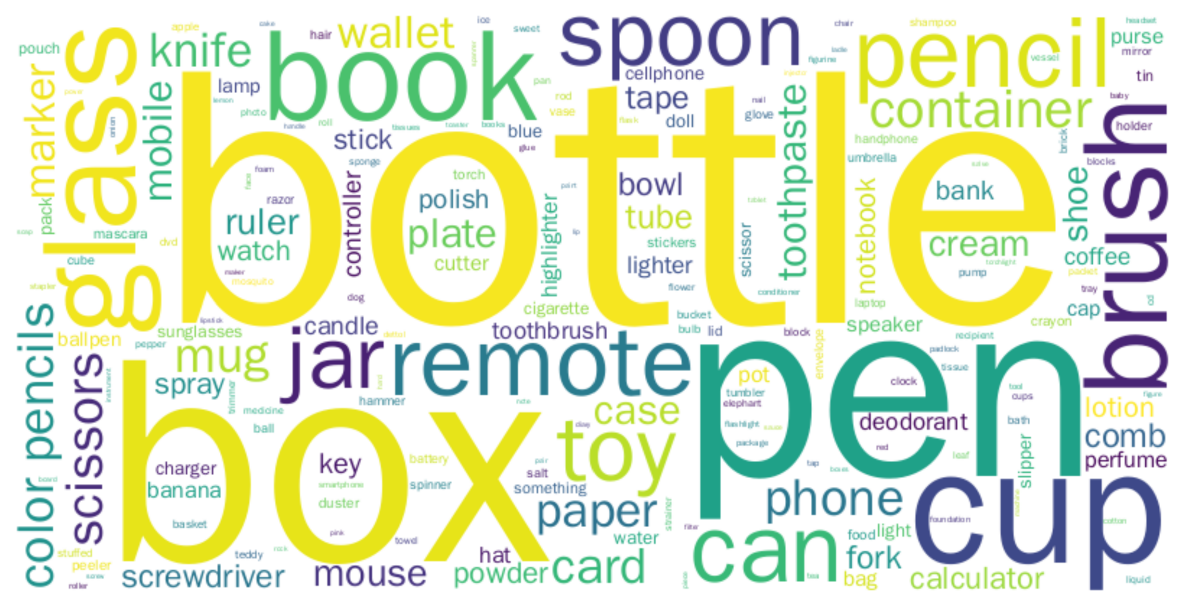}
  \caption{Object category wordcloud in the WildO2 dataset. Object categories are derived from SSv2 noun labels, with overly fine-grained descriptions merged (e.g., 'baby oil bottle' and 'perfume bottle' are grouped as 'bottle'). This reduces the number of categories from 1643 (SSv2) to 610 (Ours), followed by further refinement using VLM and manual correction.}
  \label{fig:obj_cate}
\end{figure}

For data integration purposes, Table~\ref{tab:action_label} outlines the mapping used to align labels from the Something-Something dataset with our defined action groups.
A comprehensive comparison with existing hand-object interaction datasets is presented in Table~\ref{tab:dst_comparison}. 
We further analyze the statistical properties of our collected objects, including the scale distribution and its relationship with action categories, as visualized in Figure~\ref{fig:scale_dist}. 
Figure~\ref{fig:obj_cate} presents the word cloud of object categories in our dataset.
\begin{longtable}{|c|p{0.8\textwidth}|}
\caption{The correspondence between the class labels of the Something-Something dataset and the action groups defined in this study.}\label{tab:action_label}  \\
\hline
\textbf{Action Group} & \textbf{Class Label} \\
\hline
\endfirsthead
\multirow{2}{*}{\textbf{Picking}} & Pretending to pick something up  \\ \cline{2-2}
& Picking something up \\ \hline
\multirow{10}{*}{\textbf{Pushing}} & Pushing something from right to left  \\ \cline{2-2}
& Pushing something from left to right  \\ \cline{2-2}
& Pushing something so that it almost falls off but doesn't  \\ \cline{2-2}
& Pushing something so that it slightly moves  \\ \cline{2-2}
& Pushing something so it spins  \\ \cline{2-2}
& Pushing something so that it falls off the table  \\ \cline{2-2}
& Pushing something off of something  \\ \cline{2-2}
& Pushing something onto something  \\ \cline{2-2}
& Pushing something with something  \\ \cline{2-2}
& Something colliding with something and both are being deflected \\ \hline
\multirow{9}{*}{\textbf{Poking}} & Poking something so that it falls over  \\ \cline{2-2}
& Poking a stack of something so the stack collapses  \\ \cline{2-2}
& Poking something so it slightly moves  \\ \cline{2-2}
& Poking something so lightly that it doesn't or almost doesn't move  \\ \cline{2-2}
& Pretending to poke something  \\ \cline{2-2}
& Poking something so that it spins around  \\ \cline{2-2}
& Poking a stack of something without the stack collapsing  \\ \cline{2-2}
& Poking a hole into something soft  \\ \cline{2-2}
& Poking a hole into some substance \\ \hline
\multirow{6}{*}{\textbf{Lifting}} & Lifting up one end of something without letting it drop down  \\ \cline{2-2}
& Lifting something up completely without letting it drop down  \\ \cline{2-2}
& Lifting up one end of something, then letting it drop down  \\ \cline{2-2}
& Lifting something up completely, then letting it drop down  \\ \cline{2-2}
& Lifting something with something on it  \\ \cline{2-2}
& Lifting a surface with something on it but not enough for it to slide down \\ \hline
\multirow{7}{*}{\textbf{Moving}} & Moving something down  \\ \cline{2-2}
& Moving something towards the camera  \\ \cline{2-2}
& Moving something away from the camera  \\ \cline{2-2}
& Moving something across a surface without it falling down  \\ \cline{2-2}
& Moving something up  \\ \cline{2-2}
& Moving something across a surface until it falls down  \\ \cline{2-2}
& Moving part of something \\ \hline
\multirow{2}{*}{\textbf{Squeezing}} & Pretending to squeeze something  \\ \cline{2-2}
& Squeezing something \\ \hline
\multirow{4}{*}{\textbf{Taking}} & Taking one of many similar things on the table  \\ \cline{2-2}
& Pretending to take something from somewhere  \\ \cline{2-2}
& Pretending to take something out of something  \\ \cline{2-2}
& Taking something from somewhere \\ \hline
\textbf{Touching} & Touching (without moving) part of something \\ \hline
\multirow{2}{*}{\textbf{Opening}} & Opening something  \\ \cline{2-2}
& Pretending to open something without actually opening it \\ \hline
\textbf{Laying} & Laying something on the table on its side, not upright \\ \hline
\multirow{2}{*}{\textbf{Turning}} & Pretending to turn something upside down  \\ \cline{2-2}
& Turning something upside down \\ \hline
\multirow{3}{*}{\textbf{Rolling}} & Rolling something on a flat surface  \\ \cline{2-2}
& Letting something roll along a flat surface  \\ \cline{2-2}
& Letting something roll down a slanted surface \\ \hline
\multirow{2}{*}{\textbf{Spinning}} & Spinning something that quickly stops spinning  \\ \cline{2-2}
& Spinning something so it continues spinning \\ \hline
\multirow{2}{*}{\textbf{Pulling}} & Pulling something from right to left  \\ \cline{2-2}
& Pulling something from left to right \\ \hline
\multirow{3}{*}{\textbf{Throwing}} & Throwing something  \\ \cline{2-2}
& Throwing something onto a surface  \\ \cline{2-2}
& Throwing something in the air and letting it fall \\ \hline
\multirow{9}{*}{\textbf{Putting}} & Putting something on a surface  \\ \cline{2-2}
& Putting something upright on the table  \\ \cline{2-2}
& Putting something on a flat surface without letting it roll  \\ \cline{2-2}
& Putting something onto a slanted surface but it doesn't glide down  \\ \cline{2-2}
& Putting something into something  \\ \cline{2-2}
& Putting something that can't roll onto a slanted surface, so it slides down  \\ \cline{2-2}
& Pretending to put something behind something  \\ \cline{2-2}
& Putting something that cannot actually stand upright upright on the table, so it falls on its side  \\ \cline{2-2}
& Putting something similar to other things that are already on the table \\ \hline
\multirow{2}{*}{\textbf{Falling}} & Something falling like a feather or paper  \\ \cline{2-2}
& Something falling like a rock \\ \hline
\textbf{Bending} & Bending something so that it deforms \\ \hline
\multirow{2}{*}{\textbf{Tipping}} & Tipping something over  \\ \cline{2-2}
& Tipping something with something in it over, so something in it falls out \\ \hline
\multirow{2}{*}{\textbf{Closing}} & Pretending to close something without actually closing it  \\ \cline{2-2}
& Closing something \\ \hline
\multirow{2}{*}{\textbf{Twisting}} & Twisting something  \\ \cline{2-2}
& Pretending or trying and failing to twist something \\ \hline
\textbf{Holding} & Holding something \\ \hline
\textbf{Piling} & Piling something up \\ \hline
\multirow{2}{*}{\textbf{Showing}} & Showing that something is empty  \\ \cline{2-2}
& Showing something to the camera \\ \hline
\textbf{Sprinkling} & Pretending to sprinkle air onto something \\ \hline
\multirow{3}{*}{\textbf{Tearing}} & Tearing something just a little bit  \\ \cline{2-2}
& Pretending to be tearing something that is not tearable  \\ \cline{2-2}
& Tearing something into two pieces \\ \hline
\textbf{Folding} & Folding something \\ \hline
\textbf{Unfolding} & Unfolding something \\ \hline
\textbf{Attaching} & Attaching something to something \\ \hline
\multirow{2}{*}{\textbf{Tilting}} & Tilting something with something on it slightly so it doesn't fall down  \\ \cline{2-2}
& Tilting something with something on it until it falls off \\ \hline
\textbf{Stacking} & Stacking number of something \\ \hline
\textbf{Covering} & Covering something with something \\ \hline
\textbf{Spreading} & Pretending to spread air onto something \\ \hline
\textbf{Dropping} & Dropping something onto something \\ \hline
\textbf{Wiping} & Pretending or failing to wipe something off of something \\ \hline
\end{longtable}

\begin{table}[h]
  \small
  \centering
  \begin{tabularx}{\textwidth}{@{}c|c|c|c|c|c|c|c@{}}
    \toprule
    \multirow{2}{*}[-0.5ex]{\centering Dataset} & \multicolumn{2}{c|}{Data Collection} 
    & \multicolumn{3}{c|}{Dataset Scale} & \multicolumn{2}{c@{}}{Dataset Anno.} \\
    \cmidrule(l){2-3} \cmidrule(l){4-6} \cmidrule(l){7-8}
    &  Environmnet & Authentic  & Obj Cate. & Obj Inst. & Clips/Frames & Act. Cate. & Contact  \\
    \midrule
    HO-3D      & lab & \cmark & 10 & 10 & 27/78k & none & \xmark \\
    \midrule
    obman     & syn. &\xmark & 8 & 2.7k & -/154K & none & \xmark \\
    \midrule
    HOI4D   & lab & \cmark &16 &800 & 4K/2.4M & 54 & \xmark \\      
    \midrule
    Oakink*   & lab/syn. & \cmark/\xmark &34/34 &100/1.7k& 793/230k & 5 & \xmark \\
    \midrule
    ContactPose & lab &\cmark &25 & 25 & 2.3k/3M & 25 & \cmark \\
    \midrule 
    GRAB     & lab &\cmark &37 & 51 & 1.3k/1.6M & 4 & \xmark  \\
    \midrule
    \textbf{\textit{Ours}} & wild & \cmark  & 610 & 4.4k & 4.4k/- & 92 & \cmark \\
    \bottomrule
  \end{tabularx}
  \caption{\textbf{Comparison with existing hand-object interaction datasets:} HO-3D~ \citep{ho3d}, obman~ \citep{obman}, HOI4D~ \citep{hoi4d}, Oakink*~ \citep{oakink}, ContactPose~ \citep{ctcpose}, GRAB~ \citep{grab}.
  "Authentic" indicates whether the hand-object interaction is genuine human behavior. 
  The Oakink* dataset comprises two parts: real data collected in laboratory settings (img set) and synthetic data (shape set). }
  \label{tab:dst_comparison}
  % \vspace{-15pt}
\end{table}

\subsection{Details of \methodname{}}
\label{sec:method_details}

\subsubsection{Implementation Details}
\textbf{Condition Injection Transformer Architecture}.
Our diffusion model and refiner network $f_{{refiner}}$ are both built upon an 8-layer, 4-head Transformer architecture. 
This architecture features a latent dimension of 512, a feed-forward network size of 1024, the GELU activation function, and a dropout rate of 0.1. 
To handle the variable number of input points for hands and objects, we designed an Adaptive Feature Selector, 
which employs a multi-head attention mechanism to aggregate features from the input point clouds into a fixed-size representation (64 for the hand and 128 for the object).\\
\textbf{Text Encoder}. To facilitate effective feature alignment and representation learning, we designed and implemented customized encoder architectures tailored to different sources of textual features.
For the 4096-dimensional features from the large language model Qwen-7B, we employed a non-linear feature adapter. 
This module, inspired by established practices in cross-modal alignment, utilizes a Multi-Layer Perceptron (MLP) incorporating GELU activations 
and Layer Normalization to project the high-dimensional features into a unified 512-dimensional latent space.
For the 768-dimensional token-level sequence features from models such as BERT and MPNet, we diverged from the conventional mean-pooling strategy. 
Instead, we implemented an attention-based pooling mechanism.\\
\textbf{Training}. The diffusion model is trained for 1000 epochs using the Adam optimizer with a learning rate of 1e-4 and a batch size of 128. 
Subsequently, the refiner network is trained using a distinct set of loss weights tailored for physical plausibility. 
During this phase, the parameters of the diffusion model are kept frozen.\\
\textbf{Refinement}. At inference time, following a single forward pass through the refiner network, we perform Test-Time Adaptation (TTA) on the generated pose. 
This process directly optimizes the pose parameters for 500 iterations with a learning rate of 1e-2. 
The loss weight coefficients used during all training and refinement stages are detailed in Table \ref{tab:loss_weights}.
\begin{table}[h!]
\centering
\caption{Weight coefficients for each loss term during the training and refinement stages. - indicates that the loss term is not applicable to the corresponding stage.}
\label{tab:loss_weights}
\begin{tabular}{lccl}
\toprule
\textbf{Parameter} & \textbf{Diffusion Model} & \textbf{Refiner Network} & \textbf{Description} \\
\midrule
$\lambda_{{simple}}$ & 1.0 & 5.0 & Base L2 loss \\
$\lambda_{{dmap}}$ & 0.1 & - & Hand-object distance map loss \\
$\lambda_{{global}}$ & 0.1 & 0.1 & Hand global pose loss \\
$\lambda_{{pene}}$ & - & 100.0 & Hand-object penetration loss \\
$\lambda_{{contact}}$ & - & 100.0 & Hand-object contact loss \\
$\lambda_{{cyc}}$ & - & 10.0 & Cycle-consistency loss \\
$\lambda_{{self}}$ & - & 10000.0 & Hand self-penetration loss \\
$\lambda_{{anatomy}}$ & - & 0.1 & Anatomical plausibility loss \\
\bottomrule
\end{tabular}
\end{table}

\subsubsection{Experiment Settings}
\textbf{Training Details.} We implement our model using Accelerate and train it on 8 NVIDIA 4090D GPUs with a batch size of 128.
We first define 17 detailed hand parts (e.g., pad, nail, knuckle for each finger, palmar, dorsal). Due to the long-tailed distribution of their contact frequencies, 
we aggregate these parts into 7 semantically coherent categories (pad of each finger, palmar and dorsal). 
The contact state for any interaction is then encoded as a 7-bit binary label, where each bit represents the contact status (1 for contact, 0 for non-contact) of one category. 
To create a balanced training set, we perform resampling based on these unique 7-bit labels. 

\textbf{Evaluation Metrics}
\begin{itemize}
  \item \textbf{Physical Plausibility.} We employ three metrics to assess the physical plausibility of the generated hand-object interactions:
    (1) Mean Per-Vertex Position Error (MPVPE, mm), which computes the average L2 distance between the predicted hand mesh $\hat{\mathbf{H}}$ and the ground truth $\mathbf{H}$; 
    (2) Penetration Depth (PD, cm), measuring the maximum depth of hand vertices penetrating the object surface; 
    (3) Penetration Volume (PV, cm$^3$), quantifying the volumetric intersection by voxelizing the object mesh and calculating the volume within the hand surface. 
  \item \textbf{Contact Accuracy.} Contact accuracy is measured by comparing the predicted contact map $\mathbf{C}_H$ with the ground truth $\mathbf{C}_H^*$ using Intersection over Union (IoU) and F1 score.
  \item \textbf{HOI Diversity.} Following  \citep{contactgen}, we assess diversity by clustering generated grasps into 20 clusters using K-means, and report the entropy of cluster assignments and the average cluster size.
  \item \textbf{Semantic Consistency.} Semantic consistency is evaluated by: 
    (1) P-FID, the Fréchet Inception Distance between point clouds of predicted and ground truth hand meshes, using a pre-trained feature extractor  \citep{nichol2022pointegenerating3dpoint}; 
    (2) VLM assisted evaluation, where rendered hand-object interactions are scored for semantic alignment with input captions (0-10 scale); 
    (3) Perceptual Score (PS), the mean rating from 10 volunteers (0-10 scale), reflecting the naturalness and semantic consistency of generated grasps.
\end{itemize}

\end{document}